\documentclass[10pt,journal,compsoc]{IEEEtran}

\usepackage[utf8]{inputenc}  
\usepackage[T1]{fontenc}
\usepackage{times}
\usepackage{epsfig}
\usepackage{graphicx}
\usepackage{subcaption}
\usepackage{comment}
\usepackage{amsmath}
\usepackage{amssymb}
\usepackage[table,x11names]{xcolor}
\usepackage{booktabs}
\usepackage{multirow}
\usepackage{setspace}
\usepackage{makecell}
\usepackage{stackengine}
\usepackage{microtype}
\usepackage{textcomp}
\usepackage{tikz}
\usepackage{anyfontsize}
\usepackage{psfrag}
\usepackage{mdframed}
\usepackage{ulem}
\usepackage[export]{adjustbox}
\usepackage{amssymb}%
\usepackage{pifont}%

 \usepackage{array} 
\newcolumntype{H}{>{\setbox0=\hbox\bgroup}c<{\egroup}@{}}

\newcommand{\sqdiamond}[1][fill=black]{\tikz [x=1.2ex,y=1.85ex,line width=.1ex,line join=round, yshift=-0.285ex] \draw  [#1]  (0,.5) -- (.5,1) -- (1,.5) -- (.5,0) -- (0,.5) -- cycle;}%
\newcommand{\greensquare}[1][fill={rgb:red,0.1725;green,0.6275;blue,0.1725},draw={rgb:red,0.1725;green,0.6275;blue,0.1725}]{\tikz [x=1.4ex,y=1.4ex,line width=.1ex,line join=round] \draw  [#1]  (0,0) -- (0,1) -- (1,1) -- (1,0) -- (0,0) -- cycle;}%
\newcommand{\bluediamond}{\sqdiamond[fill={rgb:red,0.1216;green,0.4667;blue,0.7059},draw={rgb:red,0.1216;green,0.4667;blue,0.7059}]}

\def\checkmark{\tikz\fill[scale=0.4](0,.35) -- (.25,0) -- (1,.7) -- (.25,.15) -- cycle;}

\newcommand{\rev}[1]{#1}
\newcommand{\revbox}[1]{#1}

\usepackage[pagebackref=true,breaklinks=true,colorlinks,bookmarks=false]{hyperref}

\ifCLASSOPTIONcompsoc
  \usepackage[nocompress]{cite}
\else
  \usepackage{cite}
\fi

\ifCLASSINFOpdf
\else
\fi

\begin{document}

\title{Physics-informed Guided Disentanglement in~Generative~Networks}

\author{Fabio~Pizzati, %
        Pietro~Cerri, %
        and~Raoul~de~Charette%
\IEEEcompsocitemizethanks{\IEEEcompsocthanksitem Fabio Pizzati and Raoul de Charette are with Inria (France) \protect\\
E-mail: fabio.pizzati@inria.fr, raoul.de-charette@inria.fr
\IEEEcompsocthanksitem Fabio Pizzati and Pietro Cerri are with Vislab Ambarella (Italy) \protect\\
Email: c-fpizzati@ambarella.com, pcerri@ambarella.com}%
\thanks{Manuscript received: xx}}

\markboth{}%
{Pizzati \MakeLowercase{\textit{et al.}}: Physics-informed Guided Disentanglement in Generative Networks}

\IEEEtitleabstractindextext{%
\begin{abstract}
Image-to-image translation (i2i) networks suffer from entanglement effects in presence of physics-related phenomena in target domain (such as occlusions, fog, etc), lowering altogether the translation quality, controllability and variability. 
\rev{In this paper, we propose a general framework to disentangle visual traits in target images.
Primarily,} we build upon collection of simple physics models, guiding the \rev{disentanglement} with a physical model that renders some of the target traits, and learning the remaining ones. 
Because \rev{physics} allows explicit and interpretable outputs, our physical models (optimally regressed on target) allows generating unseen scenarios in a controllable manner.
\rev{Secondarily, we show the versatility of our framework} to neural-guided disentanglement \rev{where a generative network is used in place of a physical model in case the latter is not directly accessible}.
\rev{Altogether, we introduce three strategies of disentanglement being guided from either a fully differentiable physics model, a (partially) non-differentiable physics model, or a neural network.}
The results show our disentanglement strategies dramatically increase performances qualitatively and quantitatively in several challenging scenarios for image translation. 
\end{abstract}

}

\maketitle

\IEEEdisplaynontitleabstractindextext

\IEEEpeerreviewmaketitle

\begin{figure}[pht]
	\centering
	\includegraphics[width=0.95\linewidth]{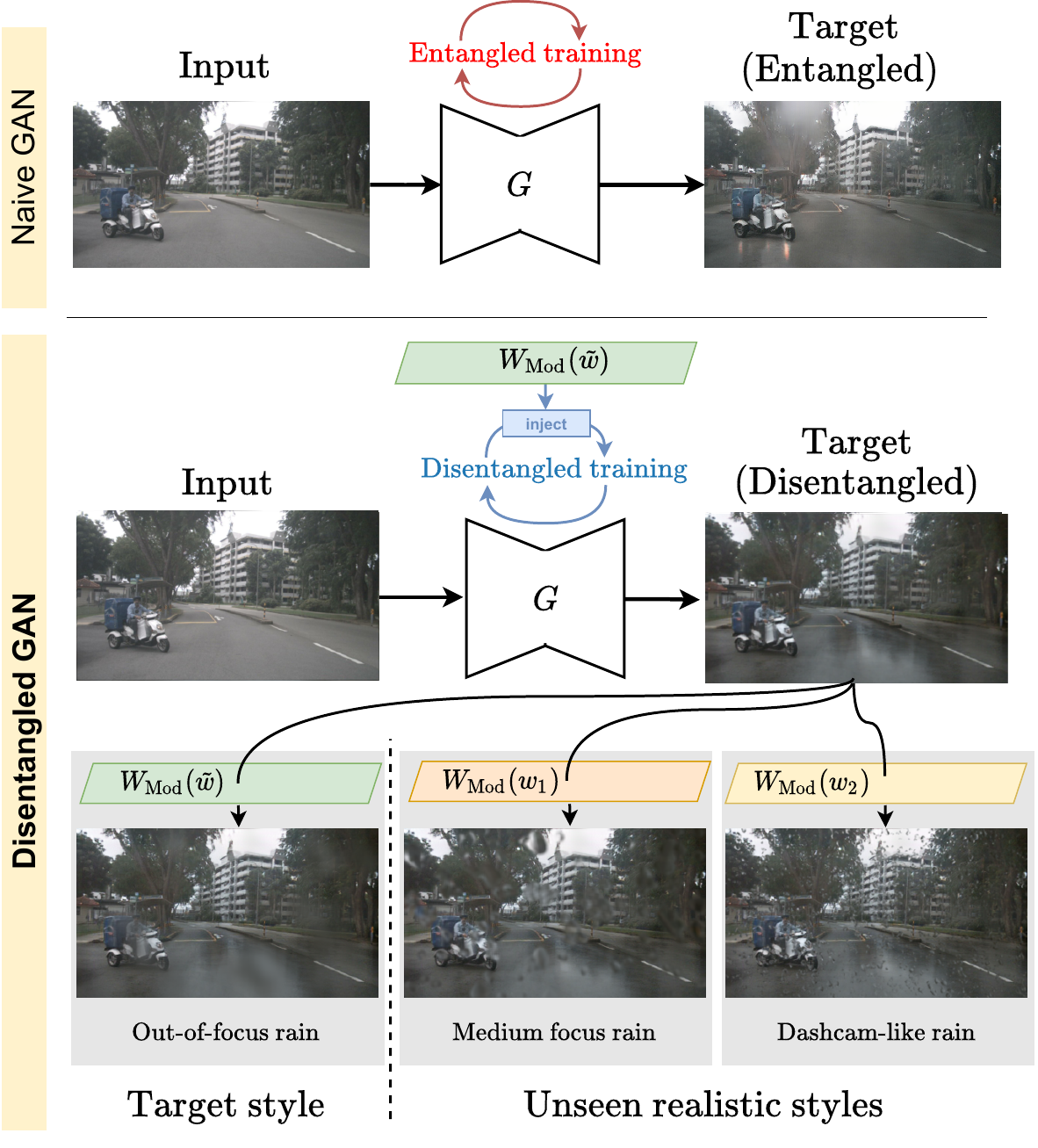}
	\caption{\textbf{Guided disentanglement.} While naive GANs generate all target scene traits at once (Target - Entangled), we learn a \textit{disentangled} version of the scene from guidance of physical model $W_\text{Mod}(.)$ with estimated physical parameters ($\tilde{w}$). Our idea is to combine physical models of well-known phenomena (as raindrops) with generative capabilities of GANs, in a complementary manner. We combine a physical model for raindrops with wetness learned by the GAN (Target - Disentangled), \textit{by only training on entangled data (i.e. rainy scene with raindrops on the lens)}. See the unrealistic raindrops in naive GANs. We instead enable the generation of target style ($\tilde{w}$) or unseen scenarios (here, $w_1$, $w_2$).
	}\label{fig:opening}
\end{figure}
\IEEEraisesectionheading{\section{Introduction}\label{sec:introduction}}

\IEEEPARstart{I}{m}age-to-image (i2i) translation GANs can learn complex style mappings in an unsupervised manner, otherwise impractical with traditional physics based rendering.
Hence, i2i GANs find great applicability in artistic style transfer, content generation, and other scenarios~\cite{zhu2017unpaired,liu2017unsupervised,isola2017image}. When coupled with domain adaptation strategies~\cite{hoffman2017cycada,li2019bidirectional,toldo2020unsupervised}, they also provide an alternative to manual labeling work for synthetic to real~\cite{bi2019deep} or challenging conditions generation~\cite{qu2019enhanced,shen2019towards,pizzati2019domain}.
However, a common pitfall of GANs is their inability to accurately learn the underlying physics of the transformation~\cite{xie2018tempogan}, often resulting in artifacts based on inaccurate mapping of source and target characteristics, which significantly impact results. 
This is the case for example when learning $\text{clear}{\mapsto}\text{rain}$, as a naive GAN translation will inevitably entangle inaccurate raindrops, as highlighted in Fig.~\ref{fig:opening}~top. On the other hand, physics-inspired models can render well-studied elements of target domain with great realism~\cite{roser2009video,alletto2019adherent,halder2019physics,tremblay2020rain}, though leaving any other appearance trait unmodified. 
For instance, in a rainy scene, models can accurately render raindrops but fail to render the complex scene wetness. 
We propose a learning-based comprehensive framework to unify generative networks and physics priors. 
We rely on a disentanglement strategy that benefits from simple physical models to learn the remaining un-modeled mapping. 
In brief, we render some of the target visual traits with a physical model and learn the un-modeled target characteristics with an i2i network. At inference, we compose them as shown in Fig.~\ref{fig:opening} to get the output benefiting from the visually pleasant outputs of GANs and the controllable characteristics of physical models. 
The peculiarity of our method is that we achieve disentanglement of modeled and learned characteristics \textit{by just using data in which they are both present simultaneously}. 
For example, we can learn to generate wet scenes \textit{without} raindrops on the lens, by only looking at rainy images \textit{with} raindrops. Our strategy deeply differs from sequential composition of i2i and physics based rendering~\cite{tremblay2020rain} which instead assume underlying independence of the two. Besides increasing image realism, our physical model-guided framework enables fine-grained control of physical parameters in rendered scenes, for increasing generated images variability regardless of the training dataset. This is beneficial for robotics applications, which require resistance to unobserved scenarios. 
A remarkable use case is vision in rainy conditions since raindrops appearances vary drastically with the camera setup. From Fig.~\ref{fig:opening} bottom, our disentanglement can be used to be resistant to dashcam-like rain even having only seen out-of-focus rain at training. Other applications we demonstrate in this paper are: vision for dirty images, foggy weather, or composite watermarks.

This research greatly extends our prior work~\cite{pizzati2020model} that focused only on occlusion disentanglement for differentiable models. \rev{We propose novel contributions that aim to address a wider spectrum of disentanglement cases, and tackle scenarios in which differentiable physical models are cumbersome to use or unavailable}. \rev{In practice}, we extend our model-guided strategy to non-differentiable models (Sec.~\ref{sec:method-genetic}), new geometry-dependent task (`Fog' in Sec.~\ref{sec:datasets}), expanding the evaluation qualitatively and quantitatively  (Sec.~\ref{sec:exp-disentanglement-physics}). 
We also conducted an extensive user study to increase the reliability of our evaluation (Secs.~\ref{sec:methodology-user-study},\ref{sec:exp-disentanglement-physics},\ref{sec:exp-validation}). 
We extend our general framework to the neural-guided disentanglement setting (Sec.~\ref{sec:method-supervised}), with new ad-hoc experiments (Sec.~\ref{sec:exp-disentanglement-neural}). 
\rev{We also extend altogether our adversarial parameters estimation (Sec.~\ref{sec:exp-validation}), ablations~(Sec.~\ref{sec:exp-ablation}) and discussion (Sec.~\ref{sec:discussion}). Finally, to encourage research in this direction, we release the code to replicate our results: \href{https://github.com/astra-vision/GuidedDisent}{https://github.com/astra-vision/GuidedDisent}.}

\section{Related works}

\subsection{Image-to-image translation}
The seminal work on image-to-image translation (i2i) using conditional GANs on paired images was conducted by Isola et al.~\cite{isola2017image}, while~\cite{wang2018pix2pixHD} exploits multi-scale architectures to generate HD results. Zhu et al.~\cite{zhu2017unpaired} propose a framework working with unpaired images introducing cycle consistency, exploited also in early work on paired multimodal image translation~\cite{zhu2017toward}. A similar idea is proposed in \cite{yi2017dualgan}.

There has been a recent trend for alternatives to cycle consistency for appearance preservation in several approaches~\cite{amodio2019travelgan,benaim2017one,fu2019geometry}, to increase focus on global image appearance and reduce it on unneeded textural preservation. In~\cite{nizan2020breaking}, they propose a cycle consistency-free multi-modal framework. Many methods also include additional priors to increase translation consistency, using objects~\cite{shen2019towards,bhattacharjee2020dunit}, instance~\cite{mo2018instagan}, geometry~\cite{wu2019transgaga,arar2020unsupervised,dell2021leveraging} or semantics~\cite{li2018semantic,ramirez2018exploiting,tang2020multi,cherian2019sem,zhu2020semantically,zhu2020sean,lin2020multimodal,ma2018exemplar,lutjens2020physicsinformed}. Other approaches learn a shared latent space using a Variational Autoencoder, as in Liu et al.~\cite{liu2017unsupervised}. 

Recently, attention-based methods were proposed, to modify partly input images while keeping domain-invariant regions unaltered \cite{mejjati2018unsupervised,ma2018gan,tang2019attention,kim2019u,Lin_2021_WACV}. Alternatively, spatial attention was exploited to drive better the adversarial training on unrealistic regions~\cite{lin2021attention}. Some methods focus instead on generating intermediate representations of source and target~\cite{gong2019dlow,lira2020ganhopper} or continuous translations~\cite{pizzati2021comogan,liu2021smoothing}. In the recent~\cite{gomez2020retrieval}, authors exploit similarity with retrieved images to increase translation quality.

\subsection{Disentanglement in i2i}
Disentangled representations of content and appearance seem to be an emerging trend to increase i2i outputs quality. Recently, Park et al.~\cite{park2020contrastive} proposed a contrastive learning based framework to disentangle content from appearance based on patches. MUNIT~\cite{huang2018multimodal}, DRIT~\cite{lee2019drit++} and TSIT~\cite{jiang2020tsit} exploit disentanglement between content and style to achieve one-to-many translations. The idea is further extended in FUNIT~\cite{liu2019few}, COCO-FUNIT~\cite{saito2020coco} and ManiFest~\cite{pizzati2021manifest} \rev{to achieve few-shot learning}, and in TUNIT~\cite{baek2021rethinking} to translate without source/target distinctions. In HiDT~\cite{anokhin2020high}, they exploit multi-scale style injection to reach translations of high definition, while~\cite{xia2020unsupervised,lin2019exploring} conditions disentanglement on domain supervision. Following different reasoning,~\cite{kondo2019flow} disentangles representations enforcing orthogonality. In~\cite{jia2020lipschitz}, they prevent semantic entanglement by using gradient regularization.

Multi-domain i2i methods~\cite{choi2018stargan,romero2019smit,anoosheh2018combogan,yang2018crossing,hui2018unsupervised,wu2019relgan,nguyen2021multi} could be also exploited for disentangling representations among different domains, at the cost of requiring annotated datasets with separated physical characteristics -- practically inapplicable for real images. 
Recent frameworks~\cite{yu2019multi,choi2020starganv2} unify multi-domain and multi-target i2i exploiting multiple disentangled representations. Some works~\cite{singh2019finegan,li2021image} detach from literature proposing hierarchical generation. In~\cite{bi2019deep}, instead, they learn separately albedo and shading, regardless of the general scene. A similar result is performed by~\cite{liu2020unsupervised}, only using unpaired images. Recently, VAE-based alternatives have also emerged~\cite{bepler2019explicitly}.

Disentangled representations could also help in physics-informed i2i tasks, such as~\cite{yang2018towards} where a fog model is exploited to dehaze images. Similarly, Gong et al.~\cite{gong2020analogical} perform fog generation exploiting paired simulated data. Even though these methods effectively learn physical transformations in a disentangled manner, they simply ignore the mapping of other domain traits.

\subsection{Physics-based generation}
Many works in literature rely on rendering to generate physics-based traits in images, for rain streaks~\cite{garg2006photorealistic,halder2019physics,tremblay2020rain,weber2015multiscale,rousseau2006realistic}, snow~\cite{barnum2010analysis}, fog~\cite{sakaridis2018semantic,halder2019physics} or others. 
In many cases, physical phenomena cause occlusion of the scene -- well studied in the literature. For instance, many models for raindrops are available, exploiting surface modeling and ray tracing~\cite{roser2009video,roser2010realistic,hao2019learning}. In~\cite{you2015adherent}, raindrop motion dynamics are also modeled. Recent works instead focus on photorealism relaxing physical accuracy constraints~\cite{porav2019can,alletto2019adherent}. A general model for lens occluders has been proposed in~\cite{gu2009removing}. Logically, it is extremely challenging to entirely simulate the appearance of scene encompassing multiple physical phenomena (for rain: rain streaks, raindrops on the lens, reflections, etc.), hence in~\cite{tremblay2020rain,musat2021multi} they also combine i2i networks and physics-based rendering. In~\cite{lengyel2021zero}, they propose to exploit night physics characteristics to perform domain adaptation.
However, this is quite different from our objective since they assume to physically model features not present in the target images. To the best of our knowledge, there is no method which unifies rendering based on physical models and i2i translations in a complementary manner. 

\begin{figure*}[t]
    \centering
    \includegraphics[width=\linewidth]{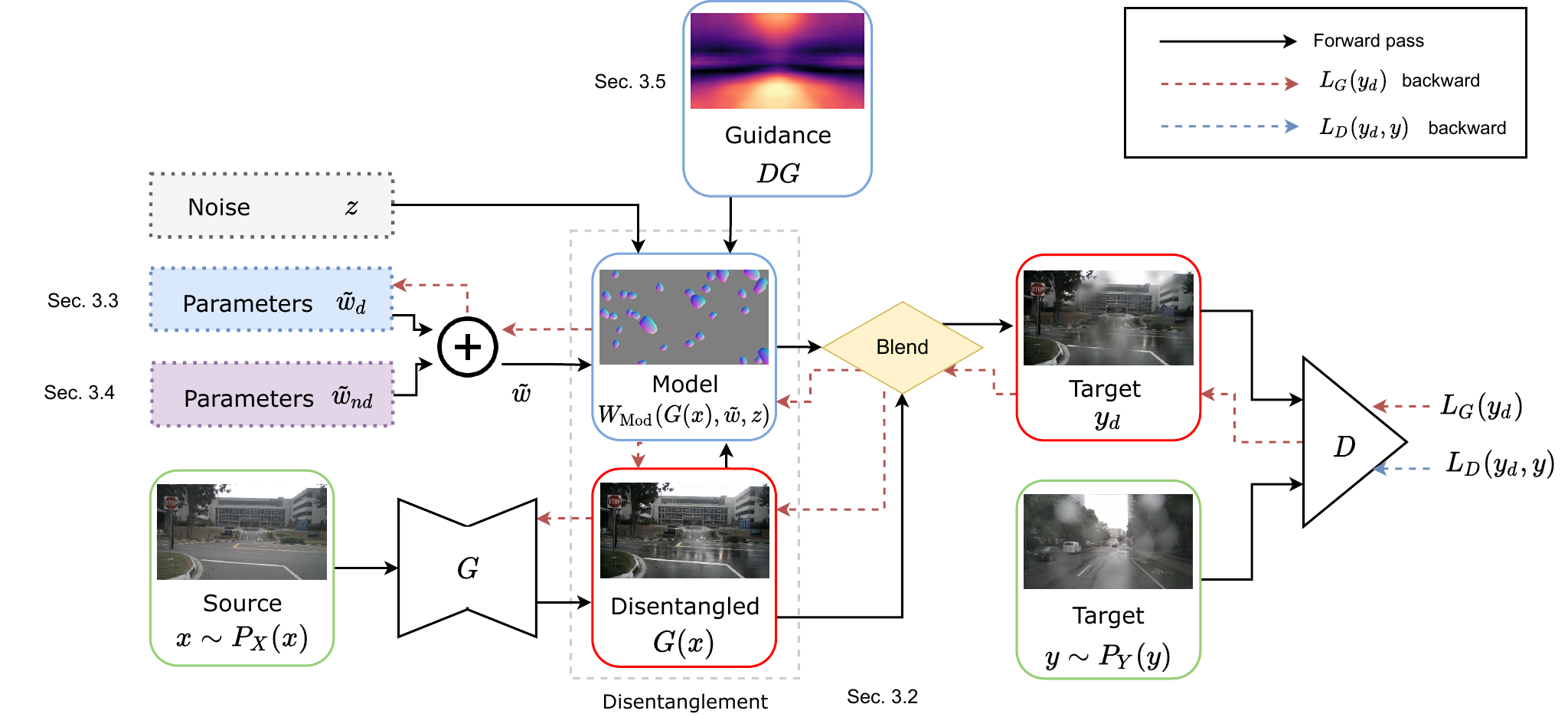}
    \caption{\textbf{Model-guided disentanglement.} Our unsupervised disentanglement process consists of applying a physical model~$W_{\text{Mod}}(.)$ to the generated image~$G(x)$. Subsequently, the composite image is forwarded to the discriminator and \rev{the GAN loss ($L_G$ or~$L_D$)} is backpropagated (dashed arrows). The model rendering depends on the estimated parameters~$\tilde{w}$, composed by differentiable ($\tilde{w}_d$) and non-differentiable ones ($\tilde{w}_{nd}$). We use a Disentanglement Guidance (DG) to avoid interfering with the gradient propagation in the learning process. Green stands for real data, red for fake ones.} 
    \label{fig:model_guided}
\end{figure*}

\section{Physical model-guided disentanglement} \label{sec:method}
\label{sec:method-unsupervised}
\label{sec:method-overview}
Standard i2i GANs solely rely on context mapping between source and target only -- which would be impractical relying only on physical rendering.
In some setups, however, the target domain encompasses some visual traits, for example adverse weather or lens occlusions, \rev{whose modeling} is well understood from physics. Hence, it may be amenable to integrate \textit{a priori} physics knowledge in the adversarial learning process. 

To formalize i2i transformations as a composition of physics and learned characteristics, we propose a setting shown in Fig.~\ref{fig:model_guided} where the GAN learns to disentangle the physically modeled traits from target~(Sec.~\ref{sec:method-disentanglement}).
Disentanglement is achieved relying on physical model-guided strategies~(Sec.~\ref{sec:method-model-disentanglement}), where we exploit as the only prior the nature of the physical trait we aim to disentangle (e.g. raindrop, dirt, fog, etc.).
Because these may have infinite variations of appearances, we estimate differentiable~(Sec.~\ref{sec:method-estimation}) and non-differentiable~(Sec.~\ref{sec:method-genetic}) target parameters of the physical model which ease disentanglement by reducing differences with target. 
Our approach boosts image quality and realism guiding model injection during training with gradient-based guidance~(Sec.~\ref{sec:method-dg}).
An extensive explanation of training strategies is in Sec.~\ref{sec:exp}.
\begin{figure*}[t]
	\centering
	\begin{subfigure}{0.6\linewidth}
		\includegraphics[width=0.95\linewidth]{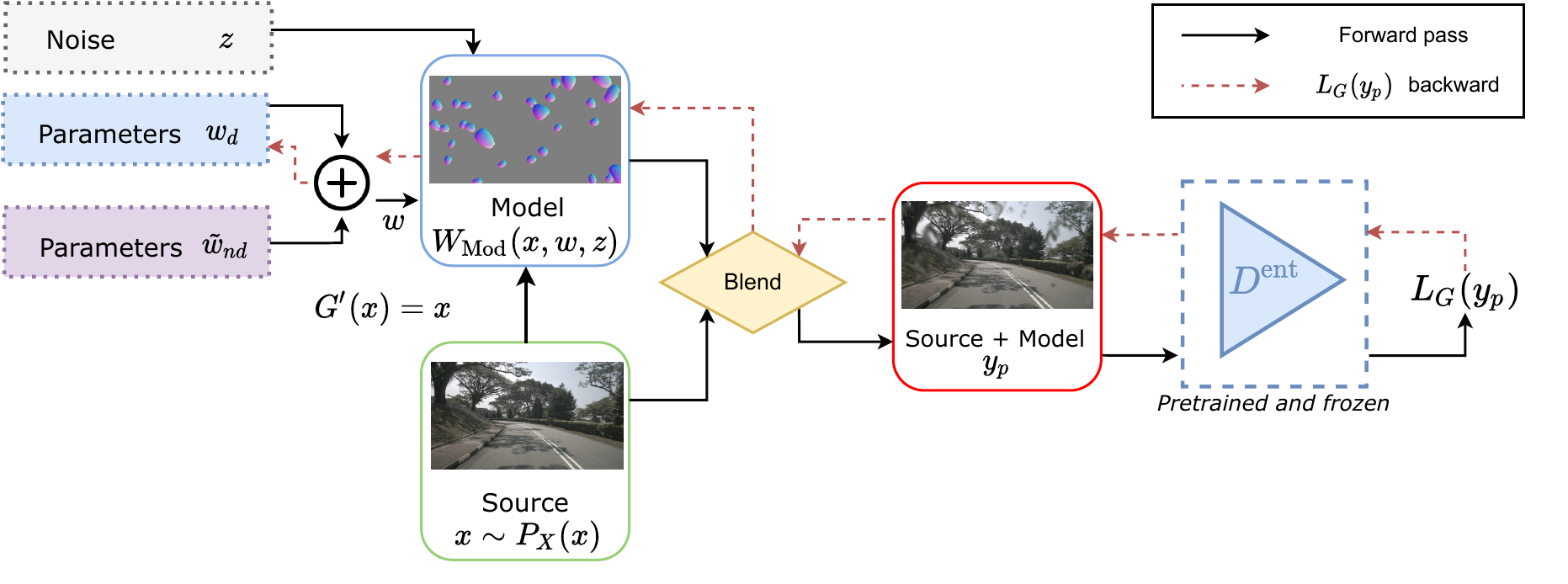}
		\caption{{Differentiable parameters.}}
		\label{fig:advest}
	\end{subfigure}\begin{subfigure}{0.4\linewidth}
	\centering
	\includegraphics[width=0.9\linewidth]{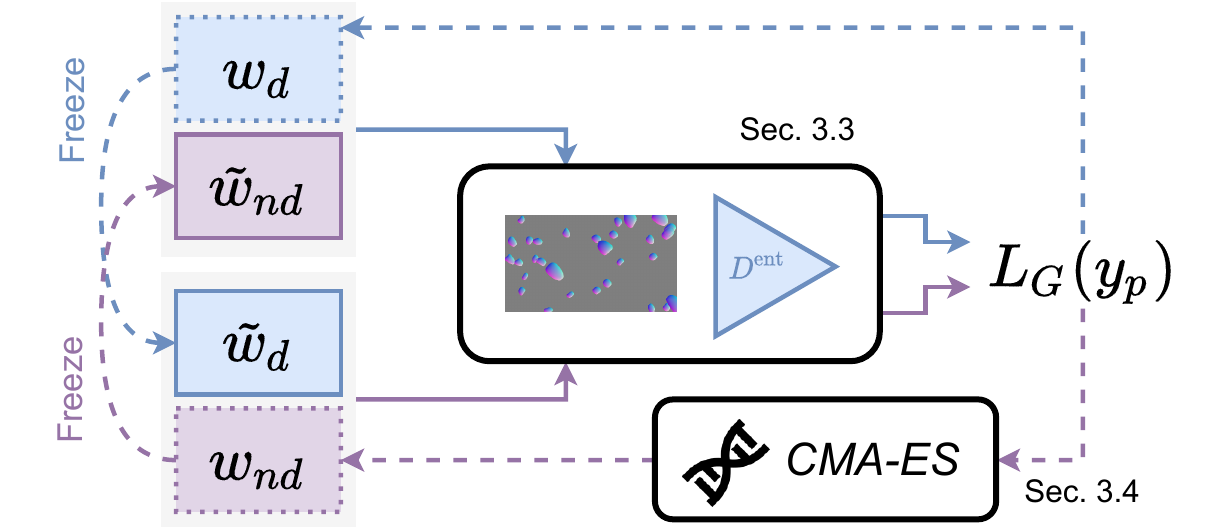}
	\caption{{Joint differentiable \textit{and} non-differentiable strategy.}}
	\label{fig:cmaes}
\end{subfigure}
\label{fig:paramestim}
\caption{\textbf{Model-guided parameters estimation.} \protect\subref{fig:advest}) We exploit a pretrained discriminator~$D^{\text{ent}}$, to calculate an adversarial loss~$L_G$ on \textit{source} data augmented with the model~$W_{\text{Mod}}$ having differentiable parameters~$w_d$. In this process, the gradient flows only in direction of the differentiable parameters. \protect\subref{fig:cmaes}) We optimize until convergence differentiable (blue) and non-differentiable (purple) parameters, alternatively reaching new minima ($\tilde{w}_d$ and~$\tilde{w}_{nd}$) used during optimization of the other parameter set. While differentiable parameters are regressed (Sec.~\ref{sec:method-estimation}), non-differentiable ones require black-box genetic optimization (Sec.~\ref{sec:method-genetic}), here CMA-ES~\cite{hansen2003reducing}.}
\end{figure*}

\subsection{Adversarial disentanglement}
\label{sec:method-disentanglement}

In image-to-image translation we aim to learn a transformation between a source~$X$ and a target~$Y$, thus mapping~$X\mapsto Y$ in an unsupervised manner. We assume that~$Y$ appearance is partly characterized by a well-identified phenomenon such as occlusions on the lens (e.g. rain, dirt) or weather phenomena (e.g. fog). Hence, we propose a sub-domain decomposition (as in ~\cite{pizzati2019domain}) of~$Y = \{Y_{W}, Y_{T}\}$, separating the identified traits ($Y_{W}$) from the other ones ($Y_{T}$). We assume this only on target, so~$X=\{X_{T}\}$.
In adversarial learning, the task of the generator is to approximate the probability distributions~$P_X$ and~$P_Y$ associated with the problem domains, such as 
\begin{equation}
    \centering
	\begin{split}
	\forall x\in X,x\sim P_X(x),\\
	\forall y\in Y,y\sim P_Y(y).
	\end{split}
\end{equation}
For explaining the intuition, we assume that the traits identifiable in this manner are independent from the recorded scene. For instance, physical properties of raindrops on a lens (such as thickness or position) do not change with the scene, as it happens also with fog, where visual effects are only depth-dependent. Therefore,~$Y_{W}$ is fairly independent from~$Y_{T}$, hence we formalize~$P_Y$ as a joint probability distribution with independent marginals, such as 
\begin{equation}
	P_Y(y) = P_{Y_W, Y_T}(y_W, y_T) = P_{Y_W}(y_W)P_{Y_T}(y_T).
	\label{eq:joint}
\end{equation}
Intuitively, approximating one of the marginals with \textit{a priori} knowledge will force the GAN to learn the other one in a disentangled manner. During training, this translates into injecting features belonging to~$Y_{W}$ before forwarding the images to the discriminator, which will provide feedback on the general realism of the image. 

Formally, we modify a LSGAN~\cite{mao2017least} training, which enforces adversarial learning minimizing

\begin{equation}
\begin{split}
y_{\text{d}} &= G(x), \\
L_{\text{gen}} = L_G(y_d) &= \mathbb{E}_{x~\sim P_X(x)}[(D(y_d) - 1) ^ 2], \\
L_{\text{disc}} = L_D(y_d, y) &= \mathbb{E}_{x~\sim P_X(x)}[D(y_d) ^ 2] + \\
&+ \mathbb{E}_{y~\sim P_Y(y)}[(D(y) - 1) ^ 2],
\end{split}
\end{equation}
\noindent{}where~$L_{\text{gen}}$ and~$L_{\text{disc}}$ are tasks of generator~$G$ and discriminator~$D$, respectively. We instead learn a disentangled mapping injecting physically modeled traits~$W_{\text{Mod}}(.)$ on translated images. We newly define~$y_{\text{d}}$ as the disentangled composition of translated scene~$G(x)$ and~$W_{\text{Mod}}(.)$, hence
\begin{equation}
\label{eq:gan_dis_objective}
    \begin{split}
    y_{\text{d}} &= \alpha_{w} G(x) + (1 - \alpha_{w})W_{\text{Mod}}(.)\,. \\
    \end{split}
\end{equation}
We define as~$\alpha_w$ a pixel-wise measure of blending between modeled and learned scene traits. Pixels which depend only on~$W_{\text{Mod}}(.)$ (as opaque occlusions) will show~$\alpha_{w} = 1$ while others (e.g. transparent ones) will have~$\alpha_{w} < 1$.

\subsection{Physics models as guidance}\label{sec:method-model-disentanglement}

One can easily obtain physical model (i.e. $W_{\text{Mod}}$) from existing literature -- typically to render visual traits like drops, fog, or else. 
Injecting such physical models in our guided-GAN enables disentanglement and learning of visual traits \textit{not} rendered by physical models, like wet materials for rain models~\cite{halder2019physics}, clouds in the sky for fog models~\cite{sakaridis2018semantic}, etc. 

However, these models often have extremely variable appearance depending on their physical parameters~$w$ so we propose adversarial-based strategies to regress optimal~$\tilde{w}$ mimicking the target dataset appearance. 
This is in fact needed for disentangled training where we assume modeled traits to resemble target ones. 
Other parameters are of stochastic nature (e.g. drop positions on the image) and are encoded as noise~$z$ regulating random characteristics.
Additionally, some models appearance -- like refractive occlusions -- varies with the underlying scene\footnote{In Sec.~\ref{sec:discussion}, we explain how~$W_{\text{Mod}}$ depending of~$s$ is not violating the independence assumption of Eq.~\ref{eq:joint}, and evaluate its effect in Sec.~\ref{sec:exp-ablation-complexity}.}~$s$, so we write~$W_{\text{Mod}}(.) = W_{\text{Mod}}(s, w, z)$, with~$s = G(x)$. 
Following our pipeline in Fig.~\ref{fig:model_guided}, if~$\tilde{w}$ properly estimates \textit{target} physical parameters,~$W_{\text{Mod}}(s, \tilde{w}, z)$ estimates marginal~$P_{Y_W}(y_W)$ which again enables disentanglement.

During inference instead,~$w$ and~$z$ can be arbitrarily varied, greatly increasing generation variability while still obtaining a realistic target scene rendering. 
In the following, we describe our adversarial parameter estimation strategy, while distinguishing differentiable ($w_d$) and non-differentiable ($w_{nd}$) parameters, such that~$w = \{w_d, w_{nd}\}$.

\subsection{Differentiable parameters estimation}
\label{sec:method-estimation}

To estimate the target optimized derivable parameters~$\tilde{w}_d$, we exploit an adversarial-based strategy benefiting from entanglement in naive trainings. We consider a naive baseline trained on~${\text{source}\mapsto\text{target}}$ mapping, where target entangles two sub-domains as specified in Sec.~\ref{sec:method-disentanglement}. We refer to generator and discriminator trained in this way as \textit{entangled} generator and discriminator, respectively. The entangled discriminator~$D^\text{ent}$ successfully learns to distinguish fake target images. \rev{This results in being able to discriminate~$P_X=P_{X_T}$ from~$P_Y=P_{Y_T}(y_T)P_{Y_W}(y_W)$.}
Considering a simplified scenario where~$P_{Y_T}$ is arbitrarily confused with the source domain, such that~$P_{Y_T} = P_{X_T}$, regressing~$w_d$ is the only way to minimize the domain shift.
In other words, considering the derivable model parametrized by~$w_d$, the above domain confusion prevents any changes in the scene. To minimize differences between source and target the network is left with updating the injected physical model appearance, ultimately regressing~$w_d$. Fig.~\ref{fig:advest} shows our differentiable parameter pipeline.
From a training perspective, we first pretrain an i2i baseline (e.g. MUNIT~\cite{huang2018multimodal}), learning a ~$X\mapsto{}Y$ mapping with an entangled generator $G^\text{ent}$ and discriminator $D^\text{ent}$. We then freeze $D^\text{ent}$ and use it to solve
\begin{equation}
\begin{gathered}
\label{eq:param-optim}
    y_p = \alpha_w x + (1 - \alpha_w)W_{\text{Mod}}(x, w, z), \min_{w_d} L_G(y_p)\,,\\
\end{gathered}
\end{equation}

\noindent{}backpropagating the GAN loss through the differentiable model.
Since many models may encompass pixelwise transparency, often the blending mask~$\alpha_w$ is~$\alpha_w = \alpha_w(w, z)$.
\noindent{}Please not this is \textit{not} a traditional adversarial training, since freezing the discriminator is mandatory to preserve the previously learned target domain appearance during the estimation process. After convergence, we extract the optimal parameter set~$\tilde{w}_d$. Alternatively,~$\tilde{w}_d$ could be manually tuned by an operator, at the cost of menial work and inaccuracy, possibly leading to errors in the disentanglement. 

From Fig.~\ref{fig:advest}, notice that the gradient flows only through differentiable parameters ($w_d$). We now detail our strategy to optimize jointly inevitable non-differentiable parameters~($w_{nd}$).

\subsection{Non-differentiable parameters estimation}
\label{sec:method-genetic}
The previously described strategy only holds for differentiable parameters~$w_{d}$, 
since we use backpropagation of an adversarial loss. Nonetheless, many models include non-differentiable parameters~$w_{nd}$ that could equally impact the realism of our model~$W_{\text{Mod}}(.)$. 
For example, a model generating raindrops occlusion would include differentiable parameters like the imaging focus, but also non-differentiable ones like the shape or number of drops -- all of which significantly impact visual appearance. 
\rev{However, the sizing of non-differentiable parameters $w_{nd}$ is both complex and time-consuming (as evaluated in Sec.~\ref{sec:exp-ablation}), and incorrect sizing is likely to achieve suboptimal disentanglement.}
Manual approximation of optimal~$w_{nd}$ parameters via trial-and-error might also be cumbersome or impractical for vast search space. To circumvent this, we exploit a genetic strategy estimating~$w_{nd}$. 

In our method, non-differentiable parameters are fed to a genetic optimization strategy. The evolutionary criteria remain the same as for differentiable parameters, that is the pretrained discriminator ($D^\text{ent}$) adversarial loss. In practice, to avoid noisy updates after genetic estimation, we average adversarial loss over a fixed number of samples to reliably select a new population. After convergence, we extract the optimal parameter set~$\tilde{w}_{nd}$.
In our experiments, we use CMA-ES~\cite{hansen2003reducing} as evolutionary strategy, but the proposed pipeline is extensible to any other genetic algorithm.
\\

\subsection{Disentanglement guidance}
\label{sec:method-dg}
It is worth noting that too sparse injection of model $W_{\text{Mod}}(.)$ negatively impacts disentanglement because the guided-GAN will entangle similar physical traits to fool the discriminator, while injecting too much of $W_{\text{Mod}}(.)$ will prevent the discovery of the disentangled target. 
Spatially, we observe that regions that do not differ from source to target are most frequently impacted by entanglement. 
This is because the discriminator naturally provides less reliable predictions due to the local source-target similarities, which leads the generator to produce artifacts resembling target physical characteristics to fool the discriminator, eventually leading to unwanted entanglement. 
In rainy scenes this happens for trees or buildings, \rev{whose appearance little varies} if dry or wet, whereas ground or road exhibit puddles which are strong rainy cues.

To balance the injection of $W_{\text{Mod}}(.)$, we guide disentanglement by injecting $W_{\text{Mod}}(.)$ only \rev{on low domain shift} areas, pushing the guided-GAN to learn the disentangled mapping of the scene. 
Specifically, we learn a Disentanglement Guidance (DG) dataset-wise by averaging the GradCAM~\cite{selvaraju2017grad} feedback on the source dataset, relying on the discriminator~$D^\text{ent}$ gradient on \textit{fake} classification. 
Areas with high domain shift will be easily identified as \textit{fake}, while others will impact less on the prediction. To take into account different resolutions, we evaluate GradCAM for all the discriminator layers. Formally, we use LSGAN to obtain
\begin{equation}
\text{DG} = \mathbb{E}_{x~\sim P_X(x)}[\mathbb{E}_{l \in L}[\text{GradCAM}_l(D^{\text{ent}}(x))]]\,,
\label{eq:method-dg}
\end{equation}
with~$L$ being the discriminator layers. At training, we inject models only on pixels~$(u,v)$ where~${\text{DG}_{u,v} < \gamma}$, with~${\gamma \in [0, 1]}$ as hyperparameter. In~Sec.~\ref{sec:exp-ablation-dg} we visually assess the effect of DG.

\begin{figure}[t]
    \centering
    \includegraphics[width=\linewidth]{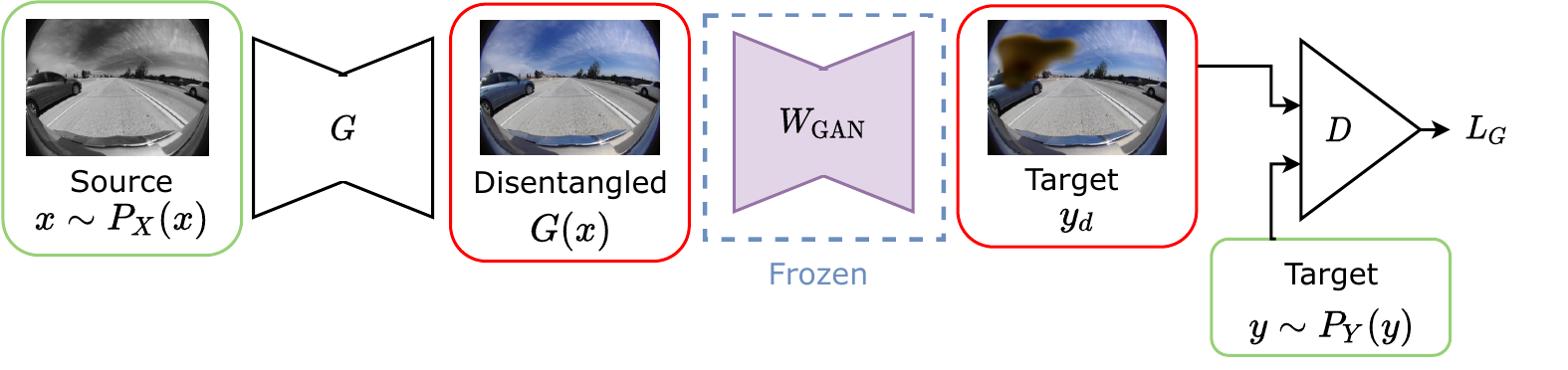}
    \caption{\textbf{Neural-guided disentanglement.} We exploit here a separate frozen GAN ($W_{\text{GAN}}$) which renders specific target traits (here, dirt) on generator~$G$ output images before forwarding them to the discriminator~$D$. We do not show gradient propagation for simplicity.}
    \label{fig:gan_guided}
\end{figure}

\subsection{Training strategy} 
\label{sec:method-training}
For models having differentiable \textit{and} non-differentiable parameters we employ a joint optimization shown in Fig.~\ref{fig:cmaes}. We first initialize a set of parameters~$w$, then alternatively use our strategy for differentiable parameters estimation~$w_{d}$~(Sec.~\ref{sec:method-estimation}) and the genetic strategy for non differentiable ones~$w_{nd}$ (Sec.~\ref{sec:method-genetic}). 
Notice that the alternation of optimized parameters prevents divergence due to simultaneous optimization. We apply updates until optimum, reaching the two sets of target style parameters, $\tilde{w} = \{\tilde{w}_{d}, \tilde{w}_{nd}\}$. The complete training strategy for model-guided disentanglement is in Sec.~\ref{sec:meth-training}.

\section{Neural-guided disentanglement}
\label{sec:method-supervised}
For some visual traits, a physical model may not be immediately available so we consider also the case in which the guidance is provided by a neural model, learned separately.
Referring to our adversarial strategy in Sec.~\ref{sec:method-disentanglement}, we simply substitute $W_{\text{Mod}}$ with $W_{\text{GAN}}$ in Eq.~\ref{eq:gan_dis_objective}, where $W_{\text{GAN}}$ is our neural guidance -- a GAN in our experiments. Following our past explanations, assuming~$W_{\text{GAN}}$ generates specific visual traits -- may it be dirt, drop, watermark or else -- it is an approximation of the marginal~$P_{Y_W}(y_W)$. We define $\tilde{\theta}$ as the optimal set of parameters of the network to reproduce target occlusion appearance. Subsequently, processing generated images with~$W_{\text{GAN}}$ before forwarding them to the discriminator pushes the guided-GAN we aim to train in a disentangled manner (not to be confused with~$W_{\text{GAN}}$) to achieve disentanglement, as illustrated in Fig.~\ref{fig:gan_guided}, following the same reasoning as in Sec.~\ref{sec:method-disentanglement}.

Of importance here, even if~$W_{\text{GAN}}$ is trained supervisedly -- for example, from annotated pairs of images / dirt -- the disentanglement strategy is itself fully unsupervised. 
Also, referring to Eq.~\ref{eq:joint}, the guided-GAN can only achieve disentanglement and estimate~$P_{Y_T}(y_T)$ from images in~$Y$, if~$W_{\text{GAN}}$ (i.e.~$W(\cdot)$) correctly estimates~$P_{Y_W}(y_W)$.
Suppose~$W_{\text{GAN}}$ augments rain on images, it will be sensitive to the intensity as well as the appearance of drops of~$Y$. In other words, it would be possible only to recreate target-like scenes\rev{, being only able to modify parameters that do not depend from appearance, as raindrops position}. With the model-guided disentanglement strategy we could instead re-inject physical traits of arbitrary appearance, greatly increasing the generative capabilities of our guided framework. \rev{Hence, the primary goal of this pipeline shall not be seen as a competitor to model-based disentanglement, but rather as a viable alternative when a physical model is not available. We present the training strategy in Sec.~\ref{sec:meth-training}.}

\section{Experiments}
\label{sec:exp}

We evaluate our disentanglement strategies on the real datasets nuScenes~\cite{caesar2019nuscenes}, RobotCar~\cite{porav2019can}, Cityscapes~\cite{cordts2016cityscapes} and WoodScape~\cite{yogamani2019woodscape}, and on the synthetic Synthia~\cite{ros2016synthia} and Weather Cityscapes~\cite{halder2019physics}. 
Our evaluation methodology is in Sec.~\ref{sec:exp-meth} including training, tasks, user study, and model/neural guidance.

In Sec.~\ref{sec:exp-disentanglement} we extensively study the disentanglement of raindrop, dirt, composite occlusions, and fog --
on a qualitative/quantitative basis, and using proxy tasks and human judgement. 
Our method is compared against the recent DRIT~\cite{lee2019drit++}, U-GAT-IT~\cite{kim2019u}, AttentionGAN~\cite{tang2019attention}, CycleGAN~\cite{zhu2017unpaired}, and MUNIT~\cite{huang2018multimodal} frameworks. 
Opposite to the literature, our method enables disentanglement of the target domain, so we report both the disentangled translations as well as the translations with the injection of optimal target physical traits. The disentanglement is greatly visible in images presented in this section. Because physical models are readily available, we emphasize our physical model-guided strategy (Sec.~\ref{sec:method-unsupervised}) evaluated on 4 models in Sec.~\ref{sec:exp-disentanglement-physics}. Conversely, the neural-guided strategy (Sec.~\ref{sec:method-supervised}), requires rare separate neural networks for rendering traits. It is subsequently only evaluated on dirt disentanglement in Sec.~\ref{sec:exp-disentanglement-neural}, relying on DirtyGAN~\cite{uricar2019let}, for comparison purposes with the model-guided strategy. In Sec.~\ref{sec:exp-validation}, we study the accuracy of our physical model parameters estimation on the well-documented raindrop model, and finally ablate our proposal in Sec.~\ref{sec:exp-ablation}.
\begin{figure}
	\centering
	\includegraphics[width=\linewidth]{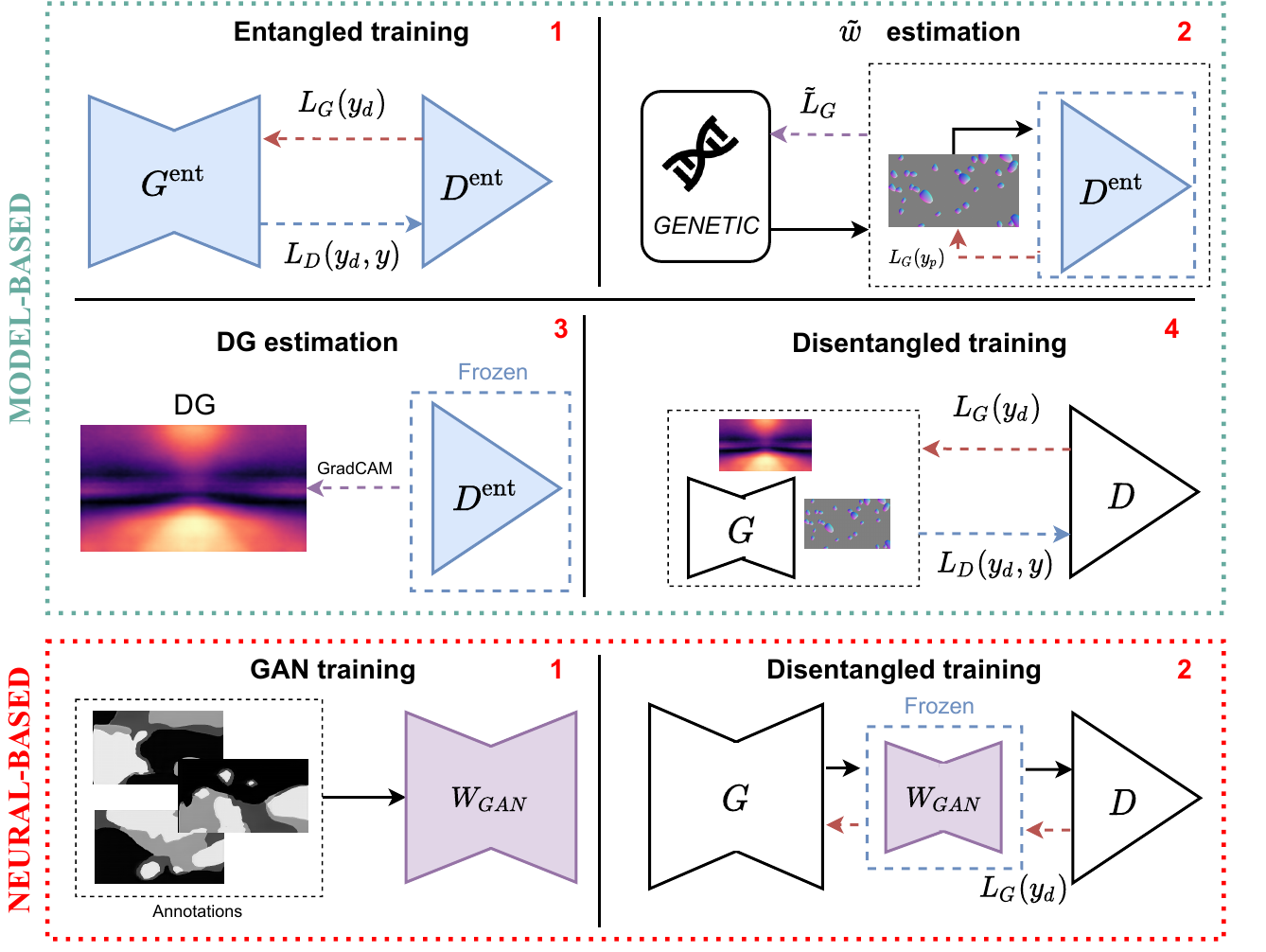}
	\caption{\textbf{Training pipelines.} For \textit{model-guided} disentanglement, we 1) train a naive i2i entangled baseline, 2) use the entangled discriminator feedback to estimate optimal parameters $\tilde{w}$ and 3) Disentanglement Guidance (DG), and finally 4) train the guided-GAN with model injection. For \textit{neural-guided} disentanglement, we 1) train a GAN ($W_{\text{GAN}}$) exploiting additional knowledge as semantics and 2) use it to inject target traits during our guided-GAN training.}
	\label{fig:training-pipelines}
\end{figure}

\noindent\textbf{Formalism.} We formalize disentangled trainings as $\mathcal{T}_{dis}$, guided either with a full physical model ($\mathcal{T}_{W_{\text{Mod}}}$), a model with only differentiable parameters ($\mathcal{T}_{W_{\text{Mod}}^{w_d}}$), or neural-guided ($\mathcal{T}_{W_{\text{GAN}}}$). 
When re-injecting physical traits, we also show their parameters in parentheses. For example,  $\mathcal{T}_{W_{\text{Mod}}}(\tilde{w})$ means model-guided disentangled output with injection of the full model estimated on target ($\tilde{w}$).

\subsection{Methodology}
\label{sec:exp-meth}

\subsubsection{Training}
\label{sec:meth-training}

Our disentangled GAN is architecture agnostic. Here, we rely on the MUNIT~\cite{huang2018multimodal} backbone for its multi-modal capabilities, and exploit LSGAN~\cite{mao2017least} for training.
Fig.~\ref{fig:training-pipelines} shows our two training pipelines.

\noindent{}For \textbf{model-guided} training (Fig.~\ref{fig:training-pipelines}, top), we leverage on a multi-step pipeline, only assuming the known nature of features to disentangle (e.g. raindrop, dirt, fog, etc.). 
First, \rev{an i2i ${\text{source}\mapsto\text{target}}$ baseline is trained} in an entangled manner, obtaining entangled 
discriminator ($D^{\text{ent}}$). 
Second, we make use of $D^{\text{ent}}$ to regress the optimal parameters $\tilde{w}$ with adversarial~(Sec.~\ref{sec:method-estimation}) and genetic~(Sec.~\ref{sec:method-genetic}) estimation. 
Third, we extract Disentanglement Guidance~(Sec.~\ref{sec:method-dg}), also using $D^{\text{ent}}$. 
Finally, we train from scratch the disentangled guided-GAN~(Sec.~\ref{sec:method-unsupervised}).\\
\noindent{}For \textbf{neural-guided} training (Fig.~\ref{fig:training-pipelines}, bottom), we use a prior-agnostic two-step pipeline. First, we train the third-party $W_{\text{GAN}}$ to render occlusions, exploiting semantic supervision in our experiments though it could realistically be replaced with self-supervision. Then, we train our disentangled guided-GAN \textit{without} any supervision.

\subsubsection{Tasks}
\label{sec:datasets}
\begin{table}
	\scriptsize
	\setlength{\tabcolsep}{0.005\linewidth}
	\renewcommand{\arraystretch}{1.0}
	\begin{tabular}{ccccccc}
		\toprule
		& \textbf{Task}  & \textbf{Entanglement} & \textbf{Datasets} & \multicolumn{3}{c}{\textbf{Guidance}} \\\hline
		& 								          & &  & \textit{Model} & $w_d$ & $w_{nd}$ \\
		\parbox[t]{4mm}{\multirow{4}{*}{\rotatebox[origin=c]{90}{\textbf{Model}}}} & $\text{clear}\mapsto\text{rain}_\text{drop}$ & Raindrop & nuScenes~\cite{caesar2019nuscenes} & Raindrop & $\sigma$ & $t$,$(s, p)$x4 \\
		&$\text{gray}\mapsto\text{color}_\text{dirt}$ & Dirt & WoodScape~\cite{yogamani2019woodscape} & Dirt & $\sigma,\alpha$ & - \\
		&$\text{synth}\mapsto\text{WCS}_\text{fog}$ & Fog & \stackunder{Synthia~\cite{ros2016synthia},}{Weather CS~\cite{halder2019physics}} & Fog & $\beta$ & -\\
		&$\text{clear}\mapsto\text{snow}_\text{cmp}$ & Composite & Synthia~\cite{ros2016synthia} & Composite  & - & -\\
		\midrule
		\parbox[t]{4mm}{\multirow{3}{*}{\rotatebox[origin=c]{90}{\textbf{Neural}}}} & & & & \multicolumn{3}{c}{\textit{Network}} \\	
		& $\text{gray}\mapsto\text{color}_\text{dirt}$  & Dirt & WoodScape~\cite{yogamani2019woodscape} & \multicolumn{3}{c}{DirtyGAN~\cite{uricar2019let}} \\
		&\\
		\bottomrule
	\end{tabular}
	\caption{\textbf{Disentanglement tasks.} For each task, we indicate the features entangled in the target domain (also, shorten as indices of task name), the datasets, and the model or neural guidance employed for disentanglement.}
	\label{tab:tasks-list}
\end{table}

Tab.~\ref{tab:tasks-list} lists the tasks evaluated and ad-hoc datasets. When referring to a task, we denote as indices the entangled features in target domain. Thus, $\text{clear}\mapsto{}\text{rain}_\text{drop}$ literally means `translation from \textit{clear} to \textit{rain} with entangled \textit{drops}'.
We later describe models used for disentanglement.\\

\begin{figure}
	\centering
	\resizebox{\linewidth}{!}{
		\setlength{\tabcolsep}{0.003\linewidth}
		\small
		\begin{tabular}{c c c c c H c H}
			& & \adjustbox{valign=m}{{\rotatebox{90}{Target}}}
			& \includegraphics[width=10em, valign=m]{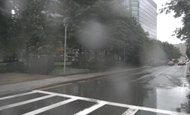}
			& \includegraphics[width=10em, valign=m]{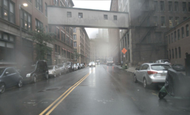}
			& \includegraphics[width=10em, valign=m]{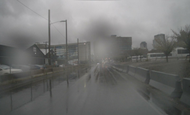}
			& \includegraphics[width=10em, valign=m]{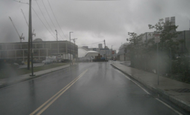}
			& \includegraphics[width=10em, valign=m]{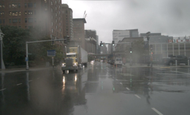}\vspace{.3em}\\
			\midrule
			& & \adjustbox{valign=m}{{\rotatebox{90}{Source}}}
			& \includegraphics[width=10em, valign=m]{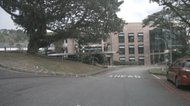}
			& \includegraphics[width=10em, valign=m]{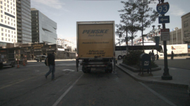}
			& \includegraphics[width=10em, valign=m]{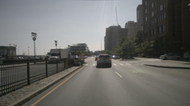}
			& \includegraphics[width=10em, valign=m]{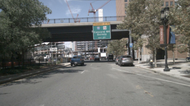}
			& \includegraphics[width=10em, valign=m]{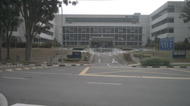}\vspace{.3em}\\
			& \adjustbox{valign=m}{{\rotatebox{90}{\cite{zhu2017unpaired}}}} & \adjustbox{valign=m}{{\rotatebox{90}{CycleGAN}}}
			& \includegraphics[width=10em, valign=m]{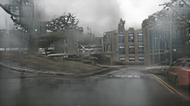}
			& \includegraphics[width=10em, valign=m]{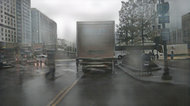}
			& \includegraphics[width=10em, valign=m]{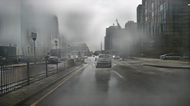}
			& \includegraphics[width=10em, valign=m]{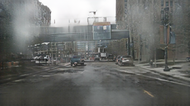}
			& \includegraphics[width=10em, valign=m]{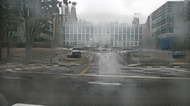}\vspace{.3em}\\
			
			& \adjustbox{valign=m}{{\rotatebox{90}{\cite{tang2019attention}}}} & \adjustbox{valign=m}{{\rotatebox{90}{Attent.GAN}}}
			& \includegraphics[width=10em, valign=m]{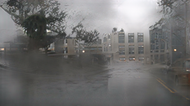}
			& \includegraphics[width=10em, valign=m]{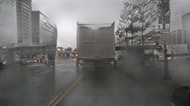}
			& \includegraphics[width=10em, valign=m]{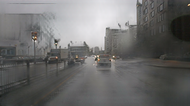}
			& \includegraphics[width=10em, valign=m]{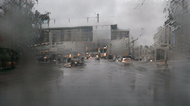}
			& \includegraphics[width=10em, valign=m]{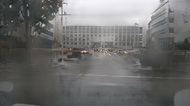}\vspace{.3em}\\
			
			& \adjustbox{valign=m}{{\rotatebox{90}{\cite{kim2019u}}}} & \adjustbox{valign=m}{{\rotatebox{90}{U-GAT-IT}}}
			& \includegraphics[width=10em, valign=m]{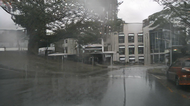}
			& \includegraphics[width=10em, valign=m]{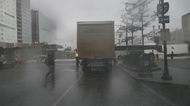}
			& \includegraphics[width=10em, valign=m]{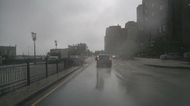}
			& \includegraphics[width=10em, valign=m]{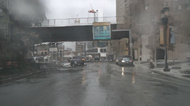}
			& \includegraphics[width=10em, valign=m]{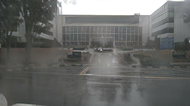}\vspace{.3em}\\
			& \adjustbox{valign=m}{{\rotatebox{90}{\cite{lee2019drit++}}}} & \adjustbox{valign=m}{{\rotatebox{90}{DRIT}}}
			& \includegraphics[width=10em, valign=m]{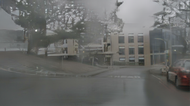}
			& \includegraphics[width=10em, valign=m]{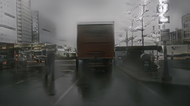}
			& \includegraphics[width=10em, valign=m]{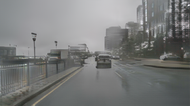}
			& \includegraphics[width=10em, valign=m]{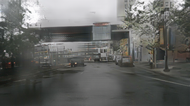}
			& \includegraphics[width=10em, valign=m]{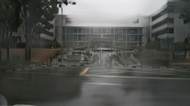}\vspace{.3em}\\
			
			& \adjustbox{valign=m}{{\rotatebox{90}{\cite{huang2018multimodal}}}} & \adjustbox{valign=m}{{\rotatebox{90}{MUNIT}}}
			& \includegraphics[width=10em, valign=m]{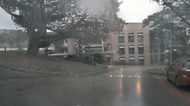}
			& \includegraphics[width=10em, valign=m]{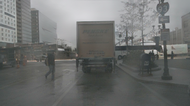}
			& \includegraphics[width=10em, valign=m]{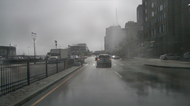}
			& \includegraphics[width=10em, valign=m]{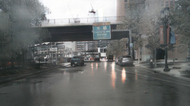}
			& \includegraphics[width=10em, valign=m]{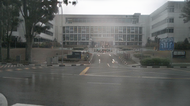}\vspace{.3em}\\
			\midrule
			\multirow{2}{*}[1em]{\rotatebox{90}{\textbf{Only differentiable}}} & \adjustbox{valign=m}{{\rotatebox{90}{\tiny{}Disentangled}}} & \adjustbox{valign=m}{{\rotatebox{90}{$\mathcal{T}_{W_{\text{Mod}}^{w_d}}$}}}
			& \includegraphics[width=10em, valign=m]{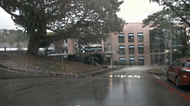}
			& \includegraphics[width=10em, valign=m]{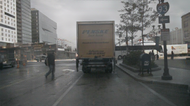}
			& \includegraphics[width=10em, valign=m]{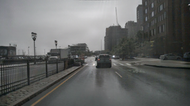}
			& \includegraphics[width=10em, valign=m]{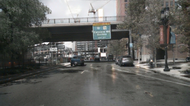}
			& \includegraphics[width=10em, valign=m]{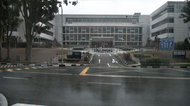}\vspace{.3em}\\
			& \adjustbox{valign=m}{{\rotatebox{90}{\tiny{}Target-style}}} & \adjustbox{valign=m}{{\rotatebox{90}{$\mathcal{T}_{W_{\text{Mod}}^{w_d}}(\tilde{w}_d)$}}}
			& \includegraphics[width=10em, valign=m]{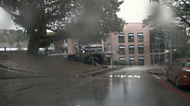}
			& \includegraphics[width=10em, valign=m]{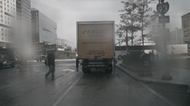}
			& \includegraphics[width=10em, valign=m]{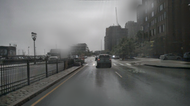}
			& \includegraphics[width=10em, valign=m]{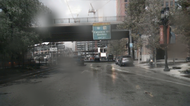}
			& \includegraphics[width=10em, valign=m]{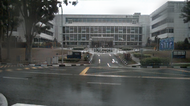}\vspace{.3em}\\
			\midrule
			\multirow{2}{*}[-1.5em]{\rotatebox{90}{\textbf{Full}}} & \adjustbox{valign=m}{{\rotatebox{90}{\tiny{}Disentangled}}} & \adjustbox{valign=m}{{\rotatebox{90}{$\mathcal{T}_{W_{\text{Mod}}^{w}}$}}}
			& \includegraphics[width=10em, valign=m]{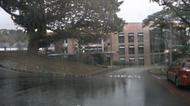}
			& \includegraphics[width=10em, valign=m]{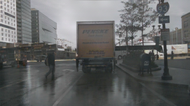}
			& \includegraphics[width=10em, valign=m]{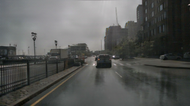}
			& \includegraphics[width=10em, valign=m]{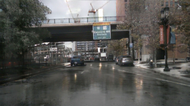}
			& \includegraphics[width=10em, valign=m]{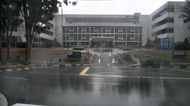}\vspace{.3em}\\
			& \adjustbox{valign=m}{{\rotatebox{90}{\tiny{}Target-style}}} & \adjustbox{valign=m}{{\rotatebox{90}{$\mathcal{T}_{W_{\text{Mod}}^{w}}(\tilde{w})$}}}
			& \includegraphics[width=10em, valign=m]{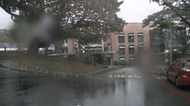}
			& \includegraphics[width=10em, valign=m]{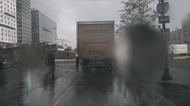}
			& \includegraphics[width=10em, valign=m]{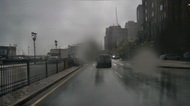}
			& \includegraphics[width=10em, valign=m]{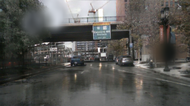}
			& \includegraphics[width=10em, valign=m]{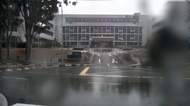}\vspace{.3em}\\
			\midrule
			\multirow{2}{*}[-1em]{\rotatebox{90}{\textbf{Unseen}}}
			& \adjustbox{valign=m}{{\rotatebox{90}{\tiny{}Dashcam-1}}} & \adjustbox{valign=m}{{\rotatebox{90}{$\mathcal{T}_{W_{\text{Mod}}^{w_d}}(w_1)$}}}
			& \includegraphics[width=10em, valign=m]{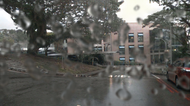}
			& \includegraphics[width=10em, valign=m]{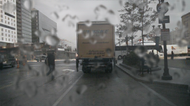}
			& \includegraphics[width=10em, valign=m]{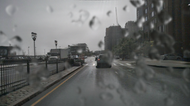}
			& \includegraphics[width=10em, valign=m]{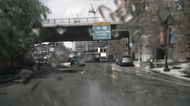}
			& \includegraphics[width=10em, valign=m]{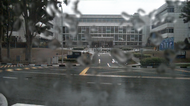}\vspace{.3em}\\
			& \adjustbox{valign=m}{{\rotatebox{90}{\tiny{}Dashcam-2}}} & \adjustbox{valign=m}{{\rotatebox{90}{$\mathcal{T}_{W_{\text{Mod}}^{w_d}}(w_2)$}}}
			& \includegraphics[width=10em, valign=m]{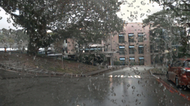}
			& \includegraphics[width=10em, valign=m]{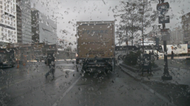}
			& \includegraphics[width=10em, valign=m]{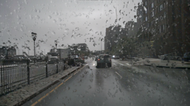}
			& \includegraphics[width=10em, valign=m]{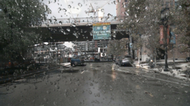}
			& \includegraphics[width=10em, valign=m]{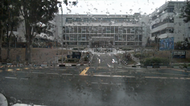}
		\end{tabular}}
		\caption{\textbf{Raindrop disentanglement on $\text{clear}\mapsto\text{rain}_\text{drop}$.}
			We compare qualitatively with the state-of-the-art on the $\text{clear}\mapsto\text{rain}_\text{drop}$ task with rain drops model-guided disentanglement. In the first row, we report samples of the target domain. Subsequently, the \textit{Source} image (2nd row), the translations by different baselines (rows 3-7) and our results (rows 8-13). Our model-guided network is able to disentangle the generation of peculiar rainy characteristics from the drops on the windshield (`Disentangled' rows) and re-injection with estimated parameters (`Target-style'). We evaluate both the differentiable-only parameter estimation (rows 8-9) and the genetic-based full estimation (rows 10-11).  We also show injection of other arbitrary parameters $w_1$, $w_2$ (last 2 rows).
		}\label{fig:qualitative-rain}
	\end{figure}

\noindent\textbf{$\text{clear}{\mapsto}\text{rain}_\text{drop}$} We exploit the recent nuScenes~\cite{caesar2019nuscenes} which includes urban driving scenes, and use metadata to build clear/rain splits obtaining 114251/29463 training and 25798/5637 testing clear/rain images.
Target rain images entangle highly unfocused drops on the windshield, which would hardly be annotated as seen in Fig.~\ref{fig:qualitative-rain}, first row.\\

\noindent\textbf{$\text{gray}{\mapsto}\text{color}_\text{dirt}$} Here, we rely on the recent fish-eye WoodScape~\cite{yogamani2019woodscape} dataset which has some images with soiling on the lens.
We separate the dataset in clean/dirty images using soiling metadata getting 5117/4873 training images and 500/500 for validation. 
Because clean/dirty splits do not encompass other domain shifts, we additionally transform \textit{clean} images to \textit{gray}.
Subsequently, we frame this as a colorization task where target \textit{color} domain entangles \textit{dirt}. 
For disentanglement, we experiment using both a physical model-guided and a neural-guided strategy.\\

\noindent\textbf{$\text{clear}{\mapsto}\text{snow}_\text{cmp}$} With Synthia~\cite{ros2016synthia} we also investigate entanglement of very different alpha-blended composites, like "Confidential" watermarks or fences. We split Synthia using metadata into clear/snow images and further augment snow target with said composite at random position. As clear/fog splits, we use 3634/3739 images for training and 901/947 for validation. To guide disentanglement, we consider a composite model, inspiring from the concept of thin occluders~\cite{garg2006photorealistic}.\\

\noindent\textbf{$\text{synth}{\mapsto}\text{WCS}_\text{fog}$} We learn here the mapping from synthetic Synthia~\cite{ros2016synthia} to the foggy version of Weather CityScapes~\cite{halder2019physics} -- a foggy-augmented Cityscapes~\cite{cordts2016cityscapes}. The goal is to learn the synthetic to real mapping, while disentangling the complex fog effect in target.  For training we use 3634/11900 and 901/2000 for validation as Synthia/WeatherCityscapes. We use a fog model to guide our network.\\
\noindent{}Note that this task differentiates from others, since target has fog of heterogeneous intensities (max. visibility 750, 375, 150 and 75m) making disentanglement significantly harder.

\subsubsection{Physical model guidance}
\label{sec:exp-meth-model}
To correctly fool the discriminator, it is crucial to choose a model that realistically resembles the entangled feature. We leverage 4 physical models, listed in Tab.~\ref{tab:tasks-list} `Model' with their differentiable~($w_d$) and non-differentiable~($w_{nd}$) parameters.

\noindent\textbf{Raindrop model.}
We extend the model of Alletto \textit{et al.}~\cite{alletto2019adherent}, which is balanced between complexity and realism. Drops are approximated by simple trigonometric functions, while we encompass also noise addition for shape variability~\cite{shadertoy}. For drops photometry, we use fixed displacement maps $(U, V)$ for coordinate mapping on both x and y axes, technically encoded as 3-channels images~\cite{alletto2019adherent}. To approximate light refraction, a drop at $(u,v)$ has its pixel $(u_i,v_i)$ mapped to
\begin{equation}
	\big(u + \text{U}(u_i,v_i)\cdot\rho, v + \text{V}(u_i,v_i)\cdot\rho\big)\,,
	\label{eq:exp-raindrops-uv}
\end{equation}
where $\rho$ is a drop-wise value representing water thickness.
Most importantly, we also model imaging focus, since it may extremely impact the rendered raindrop appearance~\cite{halimeh2009raindrop,cord2011towards,alletto2019adherent}. Hence, we use a Gaussian point spread function~\cite{pentland1987new} to blur synthetic raindrops. We implement kernel variance $\sigma$ as differentiable, while drops size ($s$), frequency ($p$), and shape ($t$) related parameters are non differentiable. We use a single shape parameter and generate 4 types of drops, with associated $p$ and $t$. 

\noindent\textbf{Dirt model.}
Here, we naively extend our raindrop model removing displacement maps as soil has no refractive behaviors. Instead, we introduce a color guidance that forces synthetic dirt to be brighter in peripherals regions, also depending on a parameter $\alpha$ which regulates occlusion maximum opacity (hence, maximum $\alpha_{w}$ value).
We also estimate $\sigma$ as aforementioned. 

\noindent\textbf{Composite occlusions model.}
We exploit the model of thin occluder proposed in~\cite{garg2006photorealistic} to render composite occlusions on images, i.e. randomly translated alpha-blended transparent images such as watermarks or fence-like grids. We assume to fully know transparency, thus no parameter is learned.

\noindent\textbf{Fog model.}
We leverage the physics model of~\cite{halder2019physics} using an input depth map. Fog thickness is regulated by a differentiable extinction coefficient $\beta$ which regulates maximum visibility.

\subsubsection{Neural guidance}
\label{sec:exp-meth-neural}
Finding appropriate neural networks to render visual traits is not trivial. Here we experiment only with Dirt, as listed in Tab.~\ref{tab:tasks-list} `Neural'.

\noindent\textbf{Dirt neural.} DirtyGAN~\cite{uricar2019let} is a GAN-based framework for opaque soiling occlusion generation. It is composed by two components, i.e. a VAE for occlusion map generation (trained using soiling semantic maps) and an i2i network conditioned on the generated map to include synthetic soiling on images.
To train DirtyGAN, we first train a VAE to learn the shape of soiling, and then proceed to train a modified CycleGAN~\cite{zhu2017unpaired} to generate realistic soiling, conditioning the soiling shape on the VAE outputs. For more details on this we refer to~\cite{uricar2019let}. 

\subsubsection{User study} \label{sec:methodology-user-study}
We also conducted a qualitative anonymous online study 
collecting answers from 56 users (22 males, 33 females, 1 non-binary) from 21 to 65 years old (mean 27.9, std. 7.6).
Each user had to evaluate 85 randomized scenes with a Likert-5 scale, providing the image looks realistic and efficiently disentangled. For ease of reading we included the results in each ad-hoc subsections \rev{(Secs.~\ref{sec:exp-disentanglement-physics}, \ref{sec:parameter-estimation})}.

\subsection{Disentanglement}
\label{sec:exp-disentanglement}
In this section, we evaluate our disentanglement strategy both using physical model-guidance (Sec.~\ref{sec:exp-disentanglement-physics}) or neural-guidance (Sec.~\ref{sec:exp-disentanglement-neural}).

\begin{figure}
	\centering
	\resizebox{\linewidth}{!}{
		\setlength{\tabcolsep}{0.003\linewidth}
		\small
		\begin{tabular}{H c c H H c c c c}
			&& \adjustbox{valign=m}{{\rotatebox{90}{Target}}}
			& \includegraphics[width=7em, valign=m]{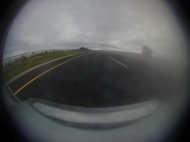}
			& \includegraphics[width=7em, valign=m]{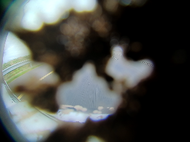}
			& \includegraphics[width=7em, valign=m]{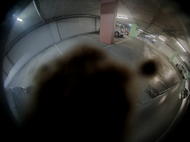}
			& \includegraphics[width=7em, valign=m]{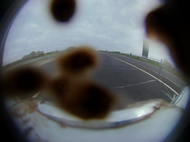}
			& \includegraphics[width=7em, valign=m]{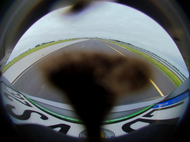}
			& \includegraphics[width=7em, valign=m]{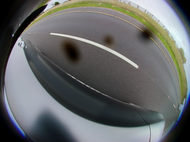}\vspace{.3em}\\
			\midrule
			&& \adjustbox{valign=m}{{\rotatebox{90}{Source}}}
			& \includegraphics[width=7em, valign=m]{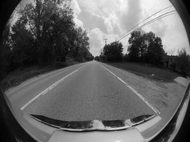}
			& \includegraphics[width=7em, valign=m]{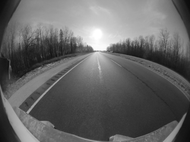}
			& \includegraphics[width=7em, valign=m]{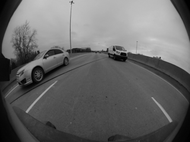}
			& \includegraphics[width=7em, valign=m]{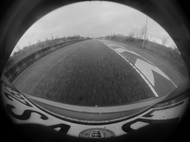}
			& \includegraphics[width=7em, valign=m]{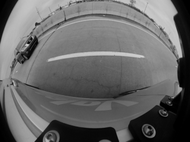}
			& \includegraphics[width=7em, valign=m]{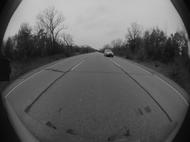}\vspace{.3em}\\
			&\adjustbox{valign=m}{{\rotatebox{90}{\cite{huang2018multimodal}}}} & \adjustbox{valign=m}{{\rotatebox{90}{MUNIT}}}
			& \includegraphics[width=7em, valign=m]{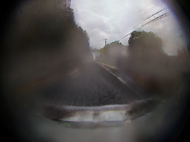}
			& \includegraphics[width=7em, valign=m]{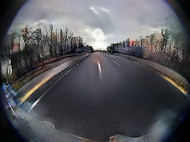}
			& \includegraphics[width=7em, valign=m]{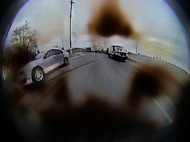}
			& \includegraphics[width=7em, valign=m]{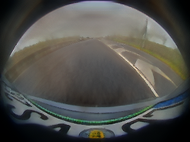}
			& \includegraphics[width=7em, valign=m]{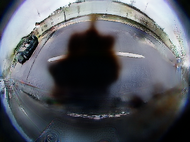}
			& \includegraphics[width=7em, valign=m]{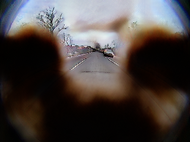}\vspace{.3em}\\
			\midrule
			\multirow{2}{*}{\rotatebox{90}{\textbf{Model-guided}}}&\adjustbox{valign=m}{{\rotatebox{90}{\tiny{}Disentangled}}} & \adjustbox{valign=m}{{\rotatebox{90}{$\mathcal{T}_{W_{\text{Mod}}^{w_d}}$}}}
			& \includegraphics[width=7em, valign=m]{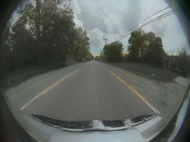}
			& \includegraphics[width=7em, valign=m]{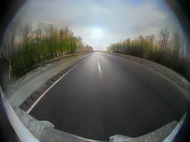}
			& \includegraphics[width=7em, valign=m]{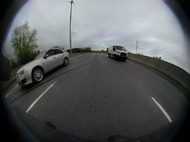}
			& \includegraphics[width=7em, valign=m]{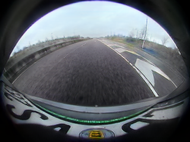}
			& \includegraphics[width=7em, valign=m]{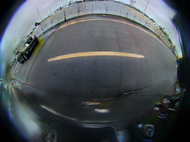}
			& \includegraphics[width=7em, valign=m]{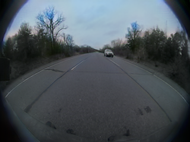}\vspace{.3em}\\
			&\adjustbox{valign=m}{{\rotatebox{90}{\tiny{}Target-style}}} & \adjustbox{valign=m}{{\rotatebox{90}{$\mathcal{T}_{W_{\text{Mod}}^{w_d}}(\tilde{w}_d)$}}}
			& \includegraphics[width=7em, valign=m]{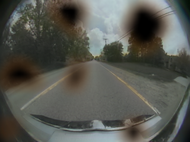}
			& \includegraphics[width=7em, valign=m]{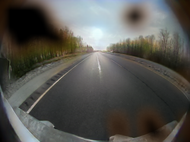}
			& \includegraphics[width=7em, valign=m]{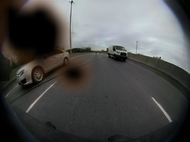}
			& \includegraphics[width=7em, valign=m]{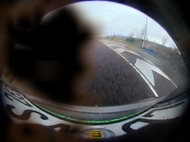}
			& \includegraphics[width=7em, valign=m]{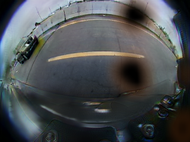}
			& \includegraphics[width=7em, valign=m]{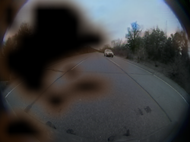}\vspace{.3em}\\

	\end{tabular}}
	\caption{\textbf{Dirt disentanglement on $\text{gray}\mapsto\text{color}_\text{dirt}$.} We compare with MUNIT \cite{huang2018multimodal} for the $\text{gray}\mapsto\text{color}_\text{dirt}$ task. Although MUNIT successfully mimics the \textit{Target} style (rows 1,3), our approach lead to a more realistic image colorization disentangling the presence of dirt (`Disentangled' row $\mathcal{T}_{W_{\text{Mod}}}$) We also use the dirt model to reproduce Target images (`Target-style' row $\mathcal{T}_{W_{\text{Mod}}^{w_d}}(\tilde{w})$). %
	} \label{fig:qualitative-dirt}
\end{figure}

\subsubsection{Physical model-guided}
\label{sec:exp-disentanglement-physics}
Referring to the 4 tasks and 4 ad-hoc models in Tab.~\ref{tab:tasks-list} `Model', we evaluate our ability to disentangle visual traits with our physical model guidance from Sec.~\ref{sec:method-unsupervised}, reporting quality, quantitative and human judgment.

Hereafter, we separate experiments on Raindrop, Dirt and Composite disentanglement from the Fog experiments, since only the former have homogeneous physical parameters~($w$) throughout the dataset\footnote{For \textit{Raindrop}, \textit{Dirt} and \textit{Composite} we consider $w_d$ and $w_{nd}$ to be dataset-wise constant. E.g. all raindrops have the same defocus blur, transparency, etc. Conversely, \textit{Fog} images have varying fog intensity.}. Since non-differentiable parameters were fairly easy to manually tune, we thoroughly experiment in the differentiable-only $\{w_d\}$ setup and compare it later on to our full $\{w_d, w_{nd}\}$ estimation (Sec.~\ref{sec:exp-validation}). 

\begin{figure}
	\centering
	\resizebox{1.0\linewidth}{!}{
	\setlength{\tabcolsep}{0.002\linewidth}
	\scriptsize
    \begin{tabular}{c c c c H c c c H c}
            &&&\multicolumn{3}{c}{\footnotesize{}\textbf{Fence}}&&\multicolumn{3}{c}{\footnotesize{}\textbf{WMK}}\\
            \cmidrule[1pt](){4-6}\cmidrule[1pt](){8-10}
			&\rotatebox{90}{\textbf{}} 
			\adjustbox{valign=m}{{\rotatebox{90}{}}} & \adjustbox{valign=m}{{\rotatebox{90}{Target}}}
			& \includegraphics[width=9em, valign=m]{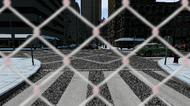}
			& \includegraphics[width=9em, valign=m]{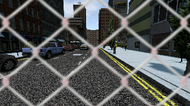}
			& \includegraphics[width=9em, valign=m]{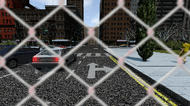}
			& \hspace{0.5em} & \includegraphics[width=9em, valign=m]{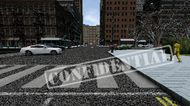}
		    & \includegraphics[width=9em, valign=m]{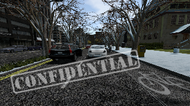}
			& \includegraphics[width=9em, valign=m]{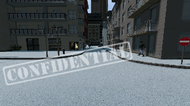}\vspace{.5em}\\ 
			
			\midrule
			&\adjustbox{valign=m}{{\rotatebox{90}{}}} & \adjustbox{valign=m}{{\rotatebox{90}{Source}}}
			& \includegraphics[width=9em, valign=m]{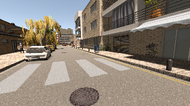}
			& \includegraphics[width=9em, valign=m]{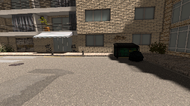}
			& \includegraphics[width=9em, valign=m]{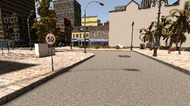}
			& \hspace{0.5em}& \includegraphics[width=9em, valign=m]{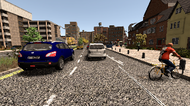}
			& \includegraphics[width=9em, valign=m]{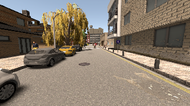}
			& \includegraphics[width=9em, valign=m]{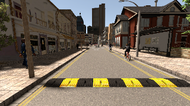}\vspace{.5em}\\ 
			
    		&\adjustbox{valign=m}{\rotatebox{90}{\cite{huang2018multimodal}}} & \adjustbox{valign=m}{{\rotatebox{90}{MUNIT}}}
			& \includegraphics[width=9em, valign=m]{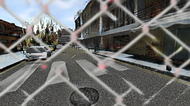}
			& \includegraphics[width=9em, valign=m]{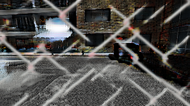}
			& \includegraphics[width=9em, valign=m]{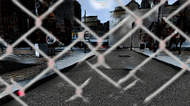}
			& \hspace{0.5em}& \includegraphics[width=9em, valign=m]{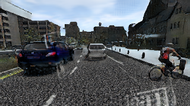}
			& \includegraphics[width=9em, valign=m]{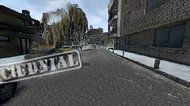}
			& \includegraphics[width=9em, valign=m]{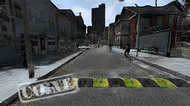}\vspace{.5em}\\ 
			
			\midrule
			\multirow{2}{*}{\rotatebox{90}{\textbf{Model-guided}}}&\adjustbox{valign=m}{{\rotatebox{90}{\tiny{}Disentangled}}} & \adjustbox{valign=m}{{\rotatebox{90}{$\mathcal{T}_{W_{\text{Mod}}^{w_d}}$}}}
			& \includegraphics[width=9em, valign=m]{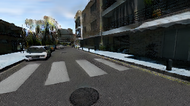}
			& \includegraphics[width=9em, valign=m]{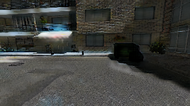}
			& \includegraphics[width=9em, valign=m]{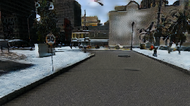}
			& \hspace{0.5em}& \includegraphics[width=9em, valign=m]{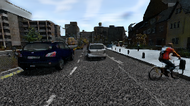}
			& \includegraphics[width=9em, valign=m]{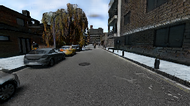}
			& \includegraphics[width=9em, valign=m]{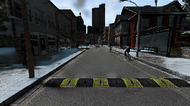}\vspace{.5em}\\ 
			
			&\adjustbox{valign=m}{{\rotatebox{90}{\tiny{}Target-style}}} & \adjustbox{valign=m}{{\rotatebox{90}{$\mathcal{T}_{W_{\text{Mod}}^{w_d}}(\tilde{w}_d)$}}}
			& \includegraphics[width=9em, valign=m]{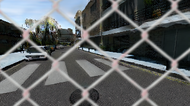}
			& \includegraphics[width=9em, valign=m]{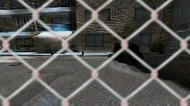}
			& \includegraphics[width=9em, valign=m]{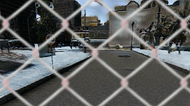}
			& \hspace{0.5em}& \includegraphics[width=9em, valign=m]{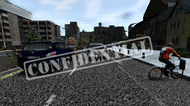}
			& \includegraphics[width=9em, valign=m]{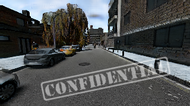}
			& \includegraphics[width=9em, valign=m]{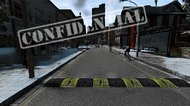}\\
		\end{tabular}}
		\caption{\textbf{Composite disentanglement on $\text{clear}\mapsto\text{snow}_\text{cmp}$.} We extend the applicability of our method to composite occlusions, that we validate in the $\text{clear}\mapsto\text{snow}_\text{cmp}$ scenario. We add a fence-like occlusion (left) and a \textit{confidential} watermark (right) to \textit{synthetic\_snow}, with random position. As expected, we encounter entanglement phenomena for MUNIT, while our model-guided network is successful in learning the disentangled appearance (`Disentangled' row $\mathcal{T}_{W_{\text{Mod}}^{w_d}}$). In our `Target-style' row $\mathcal{T}_{W_{\text{Mod}}^{w_d}}(\tilde{w}_d)$, we inject the occlusions to mimic the target style.}
\label{fig:toy}
\end{figure}

\noindent\textbf{Qualitative disentanglement.} We present different outputs for the $\text{clear}\mapsto\text{rain}_\text{drop}$ trained on nuScenes~\cite{caesar2019nuscenes}, comparing to state-of-the-art methods~\cite{lee2019drit++,kim2019u,tang2019attention,zhu2017unpaired,huang2018multimodal}~(Fig.~\ref{fig:qualitative-rain}) and for $\text{gray}\mapsto\text{color}_\text{dirt}$ and $\text{clear}\mapsto\text{snow}_\text{cmp}$ with respect to the backbone (Figs.~\ref{fig:qualitative-dirt},\ref{fig:toy}, respectively). In all cases, baselines entangle occlusions in different manners. For instance, in Fig.~\ref{fig:qualitative-rain} it is noticeable the constant position of rendered raindrops between different frameworks, as in the 1st column on the leftmost tree, which is a visible effect of entanglement and limits image variability. Also, occlusion entanglement could cause very unrealistic outputs where the structural consistency of either the scene (Fig.~\ref{fig:qualitative-dirt}) or the occlusion (Fig.~\ref{fig:toy}) is completely lost.

Referring to Figs.~\ref{fig:qualitative-rain},\ref{fig:qualitative-dirt},\ref{fig:toy}, our method is always able to produce high quality images \textit{without} occlusions (`Disentangled' rows) including typical target domain traits such as wet appearance without drops, colored image without dirt or snowy image without occlusions, respectively. 
Furthermore, we can inject occlusions with optimal estimated parameters (`Target-style' rows) to mimic target appearance which enables a fair comparison with baselines\footnote{For comparing with neural methods we set $\alpha = 1$ (cf. Sec.~\ref{sec:exp-disentanglement-neural}).}.

We also inject raindrops with arbitrary parameters to simulate \textit{unseen} dashcam-style images in Fig.~\ref{fig:qualitative-rain} (last 2 rows). The realistic results demonstrate both the quality of our disentanglement and the realism of the \textit{Raindrop} model. 
\begin{table}[t]
    \centering
    \begin{subfigure}{\linewidth}
    \resizebox{\linewidth}{!}{
        \small
        \begin{tabular}{ccccc}
        \toprule
        \textbf{Experiment} & \textbf{Network}        & \textbf{IS$\uparrow$}    & \textbf{LPIPS$\uparrow$} & \textbf{CIS$\uparrow$}   \\\midrule
    	\multirow{7}{*}{$\text{clear}\mapsto\text{rain}_\text{drop}$} & CycleGAN \cite{zhu2017unpaired} & 1.15 & 0.473 & - \\
    	& AttentionGAN \cite{tang2019attention} & 1.41 & 0.464 & - \\
    	& U-GAT-IT \cite{kim2019u} & 1.04 & 0.489 & - \\
    	& DRIT \cite{lee2019drit++} & 1.19 & 0.492 & 1.12 \\
    	& MUNIT \cite{huang2018multimodal} & 1.21 & 0.495 & 1.03\\
    	& Ours $\mathcal{T}_{W_{\text{Mod}}^{w}}(\tilde{w})$ & 1.25 & 0.502 & 1.08 \\
    	& Ours $\mathcal{T}_{W_{\text{Mod}}^{w_d}}(\tilde{w}_d)$ & \textbf{1.53} & \textbf{0.515} & \textbf{1.15} \\\midrule
        \multirow{2}{*}{$\text{gray}\mapsto\text{color}_\text{dirt}$} & MUNIT~\cite{huang2018multimodal} & 1.06 & \rev{\textbf{0.656}} & 1.08 \\
        & Ours $\mathcal{T}_{W_{\text{Mod}}^{w_d}}(\tilde{w}_d)$ & \rev{\textbf{1.25}} & 0.590 & \rev{\textbf{1.15}} \\\midrule
        $\text{clear}\mapsto\text{snow}_\text{cmp}$     & MUNIT~\cite{huang2018multimodal} & 1.26 & \textbf{0.547} & 1.11 \\ %
                                   (fence) & Ours $\mathcal{T}_{W_{\text{Mod}}^{w_d}}(\tilde{w}_d)$          & \textbf{1.31} & 0.539 & \textbf{1.19}\\\midrule
        $\text{clear}\mapsto\text{snow}_\text{cmp}$ & MUNIT~\cite{huang2018multimodal}
 & 1.17 & \textbf{0.567} & 1.01 \\ %
                                   (WMK) & Ours $\mathcal{T}_{W_{\text{Mod}}^{w_d}}(\tilde{w}_d)$            & \textbf{1.19} & 0.551 & \textbf{1.02}\\ \midrule
        \multirow{6}{*}{$\text{synth}\mapsto\text{WCS}_\text{fog}$} & CycleGAN \cite{zhu2017unpaired}
 & 1.31 & 0.384 & - \\
    	& AttentionGAN \cite{tang2019attention} & * & * & * \\
    	& U-GAT-IT \cite{kim2019u} & 1.05 & 0.406 & - \\
    	& DRIT \cite{lee2019drit++} & 1.22 & 0.424 & 1.10 \\
        & MUNIT~\cite{huang2018multimodal}
 & 1.22 & \textbf{0.429} & 1.13 \\
                           & Ours $\mathcal{T}_{W_{\text{Mod}}^{w_d}}(\tilde{w}_d)$ & \textbf{1.33} & 0.420 & \textbf{1.17}\\

        \bottomrule
    
        \end{tabular}%
        }
        \centering
        
        {
        \renewcommand{\baselinestretch}{0.5}
        \scriptsize
        \vspace{0.5em}
        \rev{* AttentionGAN converges to the identity transformation.}
	    }
        
    \caption{GAN metrics.}\label{table:ganmetrics}    \vspace{1em}
\end{subfigure}
    \begin{subfigure}{.8\linewidth}
    \centering
        \begin{subfigure}{0.9\linewidth}
    	\resizebox{\linewidth}{!}{
        \centering
        \small
        \begin{tabular}{cc}
        \toprule
			\textbf{Method} & \textbf{AP$\uparrow$} \\\midrule
			Original (from~\cite{halder2019physics}) & 18.7  \\			
			Finetuned w/ Halder \textit{et al.}~\cite{halder2019physics} & 25.6 \\		
			Finetuned w/ Model-guided $\mathcal{T}_{W_{\text{Mod}}^{w_d}}(\tilde{w}_d)$ & \textbf{27.7} \\
		\bottomrule
		\end{tabular}}\caption{Semantic segmentation on rain.}\label{table:semantic}
		\end{subfigure}
    \end{subfigure}
    \caption{\textbf{Image quality evaluation.} In (\protect{\subref{table:ganmetrics}}), we quantify GAN metrics for all tasks. While quality-aware metrics are always successfully increased, LPIPS depends on the visual complexity of the model and presence of artifacts. In (\protect{\subref{table:semantic}}), we compare our pipeline for finetuning semantic segmentation network outperforming the state-of-the-art for rain generation.}\label{table:quantitative}
\end{table}

\begin{figure}
	\centering
	\small
	\resizebox{1.0\linewidth}{!}{
	\setlength{\tabcolsep}{0.006\linewidth}
	\begin{tabular}{c c c c c c c}
			&& \adjustbox{valign=m}{{\rotatebox{90}{Target}}}
			& \includegraphics[width=12em, valign=m]{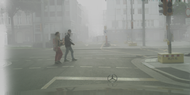}
			& \includegraphics[width=12em, valign=m]{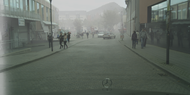}
			& \includegraphics[width=12em, valign=m]{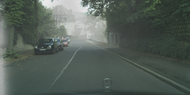}\vspace{.3em}\\
			\midrule
			&& \adjustbox{valign=m}{{\rotatebox{90}{Source}}}
			& \includegraphics[width=12em, valign=m]{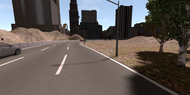}
			& \includegraphics[width=12em, valign=m]{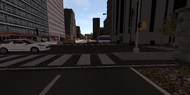}
			& \includegraphics[width=12em, valign=m]{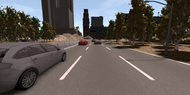}\vspace{.3em}\\
			&\adjustbox{valign=m}{{\rotatebox{90}{\cite{zhu2017unpaired}}}} & \adjustbox{valign=m}{{\rotatebox{90}{CycleGAN}}}
			& \includegraphics[width=12em, valign=m]{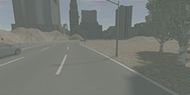}
			& \includegraphics[width=12em, valign=m]{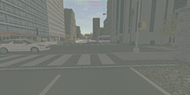}
			& \includegraphics[width=12em, valign=m]{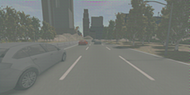}\vspace{.3em}\\
			&\adjustbox{valign=m}{{\rotatebox{90}{\cite{kim2019u}}}} & \adjustbox{valign=m}{{\rotatebox{90}{U-GAT-IT}}}
			& \includegraphics[width=12em, valign=m]{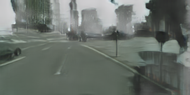}
			& \includegraphics[width=12em, valign=m]{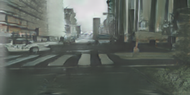}
			& \includegraphics[width=12em, valign=m]{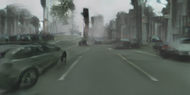}\vspace{.3em}\\
			&\adjustbox{valign=m}{{\rotatebox{90}{\cite{lee2019drit++}}}} & \adjustbox{valign=m}{{\rotatebox{90}{DRIT}}}
			& \includegraphics[width=12em, valign=m]{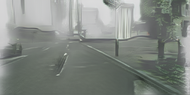}
			& \includegraphics[width=12em, valign=m]{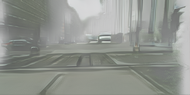}
			& \includegraphics[width=12em, valign=m]{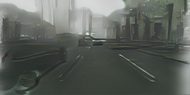}\vspace{.3em}\\
			&\adjustbox{valign=m}{{\rotatebox{90}{\cite{huang2018multimodal}}}} & \adjustbox{valign=m}{{\rotatebox{90}{MUNIT}}}
			& \includegraphics[width=12em, valign=m]{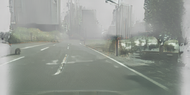}
			& \includegraphics[width=12em, valign=m]{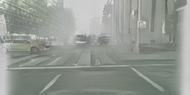}
			& \includegraphics[width=12em, valign=m]{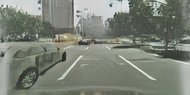}\vspace{.3em}\\
			\midrule
			\multirow{4}{*}[-4em]{\rotatebox{90}{\textbf{Model-guided}}}&\adjustbox{valign=m}{{\rotatebox{90}{\tiny{}Style $w_1$}}} & \adjustbox{valign=m}{{\rotatebox{90}{$\mathcal{T}_{W_{\text{Mod}}^{w_d}}(w_1)$}}}
			& \includegraphics[width=12em, valign=m]{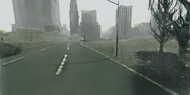}
			& \includegraphics[width=12em, valign=m]{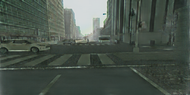}
			& \includegraphics[width=12em, valign=m]{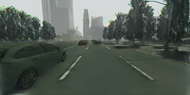}\vspace{.3em}\\
			&\adjustbox{valign=m}{{\rotatebox{90}{\tiny{}Style $w_2$}}} & \adjustbox{valign=m}{{\rotatebox{90}{$\mathcal{T}_{W_{\text{Mod}}^{w_d}}(w_2)$}}}
			& \includegraphics[width=12em, valign=m]{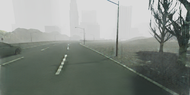}
			& \includegraphics[width=12em, valign=m]{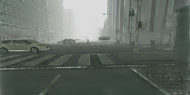}
			& \includegraphics[width=12em, valign=m]{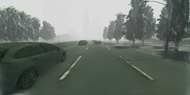}\vspace{.3em}\\
			&\adjustbox{valign=m}{{\rotatebox{90}{\tiny{}Style $w_3$}}} & \adjustbox{valign=m}{{\rotatebox{90}{ $\mathcal{T}_{W_{\text{Mod}}^{w_d}}(w_3)$}}}
			& \includegraphics[width=12em, valign=m]{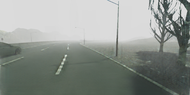}
			& \includegraphics[width=12em, valign=m]{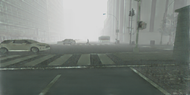}
			& \includegraphics[width=12em, valign=m]{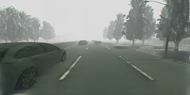}\vspace{.3em}\\
        \end{tabular}}
		\caption{\textbf{${\text{synth}\mapsto\text{WCS}_\text{fog}}$ translations.} As visible, MUNIT shows entanglement phenomena, leading to artifacts. Our model-guided disentanglement, instead, enables to generate a wide range of foggy images, with arbitrary visibility, while mantaining realism. Since the fog model $W_{\text{Mod}}$ always blocks the gradient propagation in the sky region, the network can not achieve photorealistic disentanglement but still improves the generated image quality.}\label{fig:qualitative-fog}
\end{figure}
\begin{figure*}[ht]
	\centering
	\begin{subfigure}[b]{0.333\linewidth}
		\includegraphics[width=\linewidth]{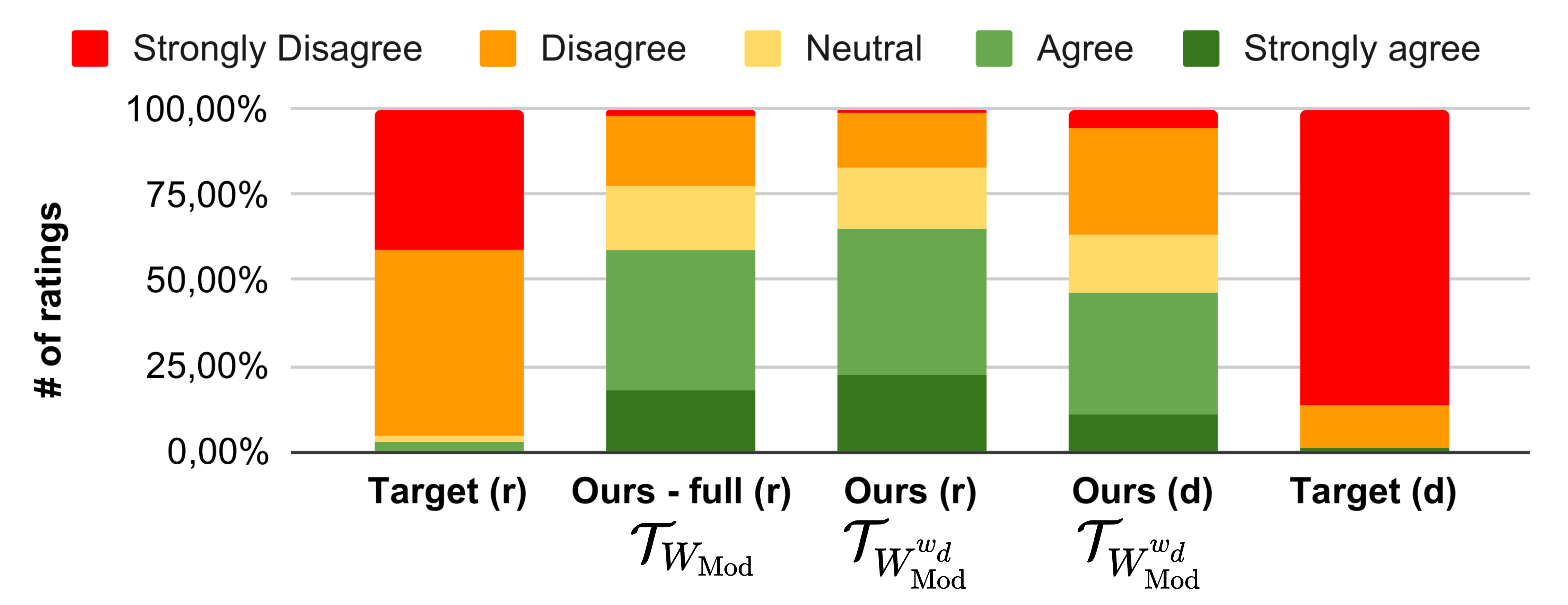}
		\caption{``The camera lens is clean''}\label{fig:userstudy-sec2}
	\end{subfigure}%
	\begin{subfigure}[b]{0.333\linewidth}
		\includegraphics[width=\linewidth]{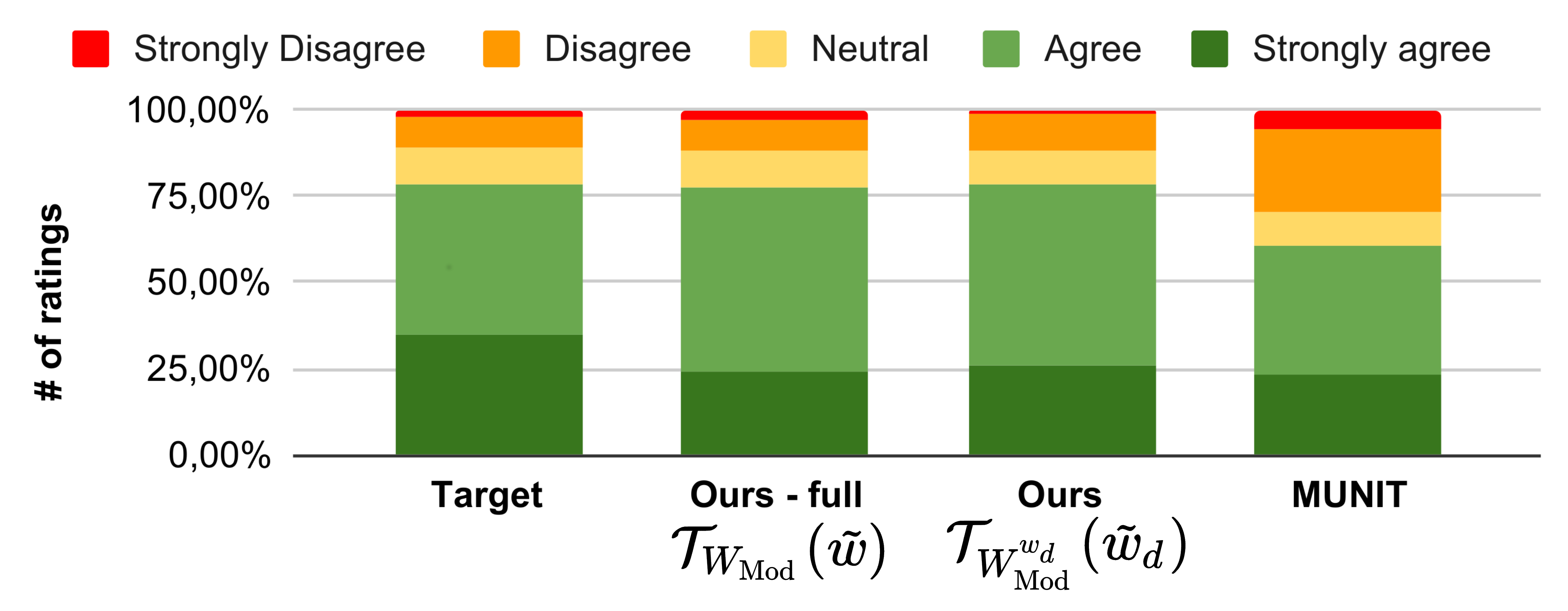}
		\caption{``The scene looks wet'' %
		}
		\label{fig:userstudy-sec5}
	\end{subfigure}%
	\begin{subfigure}[b]{0.333\linewidth}
		\includegraphics[width=\linewidth]{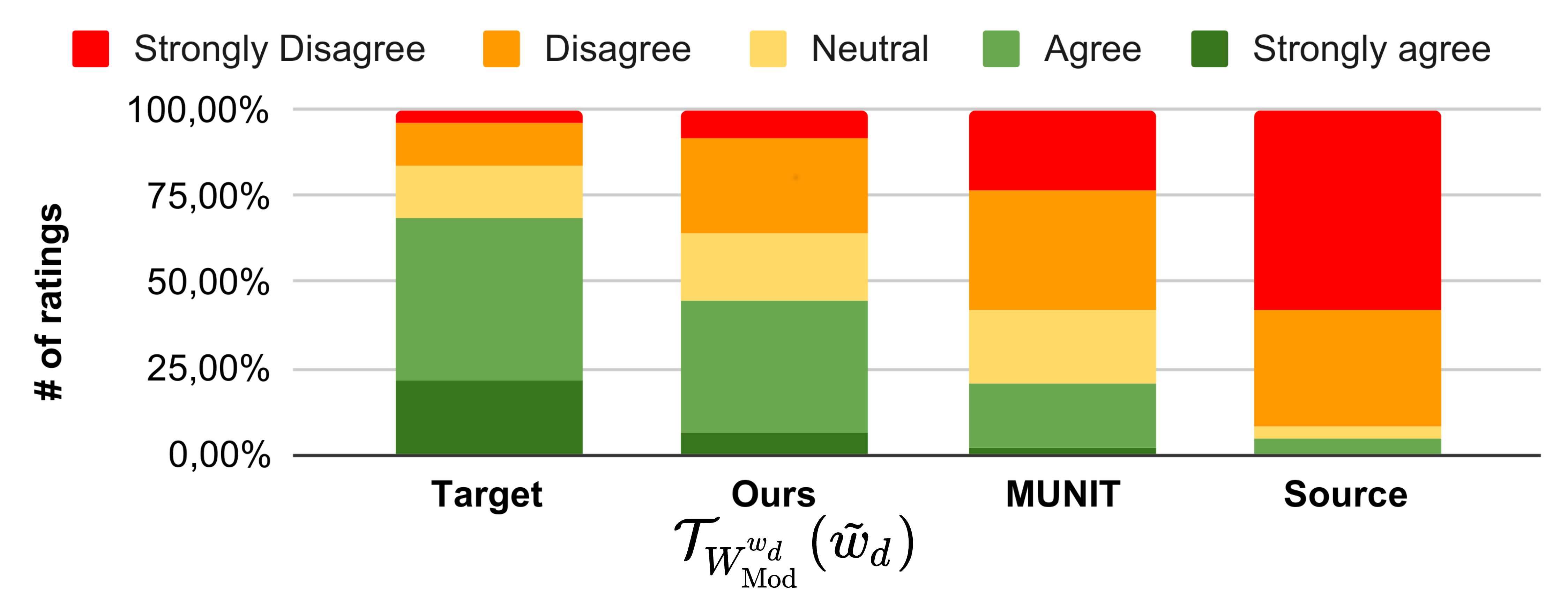}
		\caption{``The colors look natural'' %
		}\label{fig:userstudy-sec4}
	\end{subfigure}
	\caption{\textbf{Disentanglement user study.} We asked 56 users (cf. Sec.~\ref{sec:methodology-user-study}) to judge the lens cleanness (\subref{fig:userstudy-sec2}) on raindrops (r) and dirt (d), or the wetness (\subref{fig:userstudy-sec5}) or coloring (\subref{fig:userstudy-sec4}) of $\text{clear}\mapsto\text{rain}_{\text{drop}}$ and $\text{gray}\mapsto\text{color}_{\text{dirt}}$ generated scenes, respectively. Details are in the text. Our system greatly improves results following human evaluation metrics.}
\end{figure*}

\begin{figure}[ht]
	\centering
	\begin{subfigure}{0.535\linewidth}
		\resizebox{\linewidth}{!}{
			\setlength{\tabcolsep}{0.003\linewidth}
			\small
			\begin{tabular}{Hcccc}
				\toprule
				\textbf{Experiment} & \textbf{Network}        & \textbf{IS$\uparrow$}    & \textbf{LPIPS$\uparrow$} & \textbf{CIS$\uparrow$}   \\
				\midrule
				\multirow{3}{*}{$\text{gray}\mapsto\text{color}_\text{dirt}$} & MUNIT~\cite{huang2018multimodal} & 1.06 & 0.656 & 1.08 \\
				& Model-guided $\mathcal{T}_{W_{\text{Mod}}^{w_d}}(\tilde{w}_d)$ & 1.25 & 0.590 & 1.15 \\
				& Neural-guided $\mathcal{T}_{W_{\text{GAN}}}(\tilde{\theta})$ & \textbf{1.58} & \textbf{0.663} & \textbf{1.47} \\
				\bottomrule
				
			\end{tabular}}
			\caption{GAN metrics.}\label{table:ganguided-ganmetrics}\end{subfigure} \begin{subfigure}{0.455\linewidth}
			\resizebox{\linewidth}{!}{
				\setlength{\tabcolsep}{0.003\linewidth}
				\centering
				\small
				\begin{tabular}{ccc}
					\toprule
					\textbf{Network} & \textbf{SSIM$\uparrow$} & \textbf{PSNR$\uparrow$} \\\midrule
					MUNIT \cite{huang2018multimodal} & 0.414 & 13.4\\			
					Model-guided $\mathcal{T}_{W_{\text{Mod}}^{w_d}}$ & \textbf{0.755} & \textbf{20.2}\\
					Neural-guided $\mathcal{T}_{W_{\text{GAN}}}$ & 0.724 & 19.3\\\bottomrule			
				\end{tabular}}\caption{Colorization.}\label{table:ganguided-colorization}
			\end{subfigure}
			\\
			\vspace{2em}
			\begin{subfigure}{\linewidth}
				\revbox{
					\resizebox{\linewidth}{!}{
						\setlength{\tabcolsep}{0.003\linewidth}
						\small
						\begin{tabular}{c c H H H c c c c}
							&\adjustbox{valign=m}{{\rotatebox{90}{Target}}}& 
							& \includegraphics[width=7em, valign=m]{figures/dirt_images/1_FV_soil.png}
							& \includegraphics[width=7em, valign=m]{figures/dirt_images/3_soil.png}
							& \includegraphics[width=7em, valign=m]{figures/dirt_images/6_soil.png}
							& \includegraphics[width=7em, valign=m]{figures/dirt_images/99_FV_soil.png}
							& \includegraphics[width=7em, valign=m]{figures/dirt_images/393_RV_soil.png}
							& \includegraphics[width=7em, valign=m]{figures/dirt_images/263_MVL_soil.png}\vspace{.3em}\\\midrule
							&\adjustbox{valign=m}{{\rotatebox{90}{Source}}}& 
							& \includegraphics[width=7em, valign=m]{figures/dirt_images/14_image.png}
							& \includegraphics[width=7em, valign=m]{figures/dirt_images/21_image.png}
							& \includegraphics[width=7em, valign=m]{figures/dirt_images/23_image.png}
							& \includegraphics[width=7em, valign=m]{figures/dirt_images/25_image.png}
							& \includegraphics[width=7em, valign=m]{figures/dirt_images/43_image.png}
							& \includegraphics[width=7em, valign=m]{figures/dirt_images/27_image.png}\vspace{.3em}\\
							&\adjustbox{valign=m}{{\rotatebox{90}{Ground Truth}}}& 
							& \includegraphics[width=7em, valign=m]{figures/dirt_images/14_image.png}
							& \includegraphics[width=7em, valign=m]{figures/dirt_images/21_image.png}
							& \includegraphics[width=7em, valign=m]{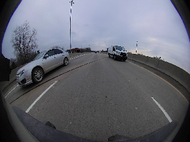}
							& \includegraphics[width=7em, valign=m]{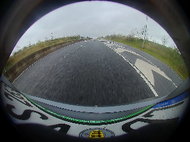}
							& \includegraphics[width=7em, valign=m]{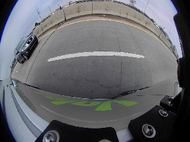}
							& \includegraphics[width=7em, valign=m]{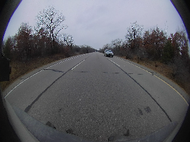}\vspace{.3em}\\
							\midrule
							\multirow{2}{*}{\rotatebox{90}{\textbf{Model-guided}}}&\adjustbox{valign=m}{{\rotatebox{90}{\small{}Disentangled}}} & \adjustbox{valign=m}{{\rotatebox{90}{$\mathcal{T}_{W_{\text{Mod}}^{w_d}}$}}}
							& \includegraphics[width=7em, valign=m]{figures/dirt_images/14_translated.png}
							& \includegraphics[width=7em, valign=m]{figures/dirt_images/21_translated.png}
							& \includegraphics[width=7em, valign=m]{figures/dirt_images/23_translated.png}
							& \includegraphics[width=7em, valign=m]{figures/dirt_images/25_translated.png}
							& \includegraphics[width=7em, valign=m]{figures/dirt_images/43_translated.png}
							& \includegraphics[width=7em, valign=m]{figures/dirt_images/27_translated.png}\vspace{.3em}\\
							&\adjustbox{valign=m}{{\rotatebox{90}{\small{}Target-style}}} & \adjustbox{valign=m}{{\rotatebox{90}{$\mathcal{T}_{W_{\text{Mod}}^{w_d}}(\tilde{w}_d)$}}}
							& \includegraphics[width=7em, valign=m]{figures/dirt_images/14_dirtymodel.png}
							& \includegraphics[width=7em, valign=m]{figures/dirt_images/21_dirtymodel.png}
							& \includegraphics[width=7em, valign=m]{figures/dirt_images/23_dirtymodel.png}
							& \includegraphics[width=7em, valign=m]{figures/dirt_images/25_dirtymodel.png}
							& \includegraphics[width=7em, valign=m]{figures/dirt_images/43_dirtymodel.png}
							& \includegraphics[width=7em, valign=m]{figures/dirt_images/27_dirtymodel.png}\vspace{.3em}\\
							\midrule
							\multirow{2}{*}{\rotatebox{90}{\textbf{Neural-guided}}}&\adjustbox{valign=m}{{\rotatebox{90}{\small{}Disentangled}}} & \adjustbox{valign=m}{{\rotatebox{90}{$\mathcal{T}_{W_{\text{GAN}}}$}}}
							& \includegraphics[width=7em, valign=m]{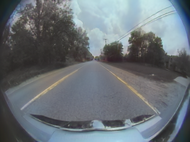}
							& \includegraphics[width=7em, valign=m]{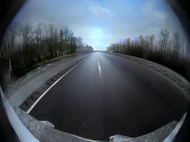}
							& \includegraphics[width=7em, valign=m]{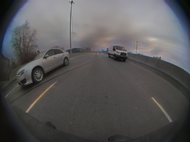}
							& \includegraphics[width=7em, valign=m]{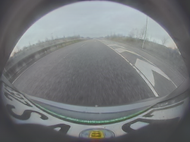}
							& \includegraphics[width=7em, valign=m]{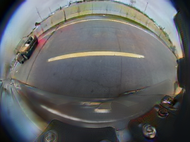}
							& \includegraphics[width=7em, valign=m]{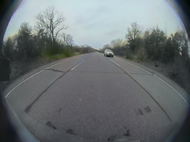}\vspace{.3em}\\
							&\adjustbox{valign=m}{{\rotatebox{90}{\small{}Target-style}}} & \adjustbox{valign=m}{{\rotatebox{90}{$\mathcal{T}_{W_{\text{GAN}}}(\tilde{\theta})$}}}
							& \includegraphics[width=7em, valign=m]{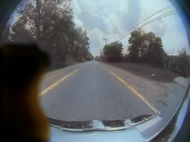}
							& \includegraphics[width=7em, valign=m]{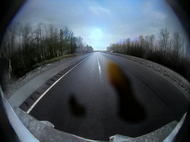}
							& \includegraphics[width=7em, valign=m]{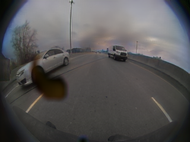}
							& \includegraphics[width=7em, valign=m]{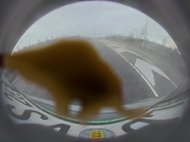}
							& \includegraphics[width=7em, valign=m]{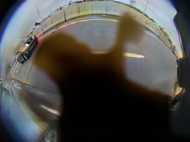}
							& \includegraphics[width=7em, valign=m]{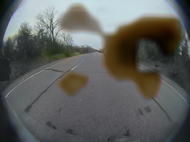}\vspace{.3em}\\

						\end{tabular}}\caption{Qualitative evaluation.}\label{table:ganguided-qual}
					}
				\end{subfigure}

				\caption{\textbf{Comparison of model- and neural- guided disentanglement on $\text{gray}\mapsto\text{color}_\text{dirt}$.} Although our neural-guided strategy excels in image quality and diversity, mostly due to the complex nature of generated dirt (\subref{table:ganguided-ganmetrics}), with model guidance we achieve more realistic image colorization (\subref{table:ganguided-colorization}). Qualitative results are coherent with metrics (\subref{table:ganguided-qual}). With both pipelines, we still outperform MUNIT~\cite{huang2018multimodal}, used as backbone.} \label{fig:ganguided}
			\end{figure}

\noindent\textbf{Quantitative disentanglement.} We use GAN metrics to quantify the quality of the learned mappings. Results are reported in Tab.~\ref{table:ganmetrics}, where Inception Score (IS)~\cite{salimans2016improved} evaluates quality and diversity against target, LPIPS distance~\cite{zhang2018unreasonable} evaluates translation diversity (thus avoiding mode-collapse), and Conditional Inception Score (CIS)~\cite{huang2018multimodal} single-image translations diversity for multi-modal baselines. 
In practice, IS is computed over all the validation set while CIS is estimated on 100 different translations of 100 random images following~\cite{huang2018multimodal}. The InceptionV3 network for Inception Scores was finetuned on the source/target classification as in~\cite{huang2018multimodal}. LPIPS distance is calculated on 1900 random pairs of 100 translations as in~\cite{huang2018multimodal}.
For fairness, we only compare `Target-style' outputs to baselines, since those are not supposed to disentangle physical traits, and can only output images resembling \textit{Target}.  \\
\noindent{}Tab.~\ref{table:ganmetrics} shows we outperform all baselines on IS/CIS, including MUNIT -- our i2i backbone.
This is due to disentanglement, since entanglement phenomena limit occlusions appearance and position variability. Even the scene translation quality is improved by disentanglement since the generator learns a simpler target domain mapping without any occlusions. 
As regards LPIPS distance, we outperform the baseline on raindrops while we rank lower on the other tasks. While IS/CIS quantify both quality and diversity, LPIPS metric is evaluating variability only thus penalizing simpler occlusion generation. For instance, our rendered dirt in Fig.~\ref{fig:qualitative-dirt} is often black while MUNIT-generated artifacts are highly variable (compare rows \textit{MUNIT} and ours $\mathcal{T}_{W_{\text{Mod}}^{w_d}}(\tilde{w}_d)$). The same happens for watermarks in Fig.~\ref{fig:toy}, where unrealistic artifacts are highly variable. For raindrops, instead, MUNIT tends to just blur images, while we benefit from the refractive capabilities of our physical model which increase LPIPS.

\noindent{}{\textit{Semantic segmentation.}}
To provide additional insights on the effectiveness of our framework and compensate for the well-known noisiness of GAN metrics~\cite{zhang2018unreasonable}, we quantify the usability of generated images for semantic segmentation in the $\text{clear}\mapsto\text{rain}_{\text{drop}}$ setup.
Therefore, we process the popular Cityscapes~\cite{cordts2016cityscapes} dataset for semantic segmentation with our best $\text{clear}\mapsto\text{rain}_\text{drop}$ model-guided training, obtaining a synthetic rainy version $\mathcal{T}_{W_{\text{Mod}}^{w_d}}(\tilde{w}_d)$ that we use for finetuning PSPNet~\cite{zhao2017pyramid}, following Halder et al.~\cite{halder2019physics}. 
Please note that this also demonstrates the generation capabilities to new scenarios of our GAN, since we use the pretrained network on nuScenes given the absence of rainy scenes in Cityscapes.
We report the mAP for the 25 rainy images with semantic labels provided by~\cite{halder2019physics} in Tab.~\ref{table:semantic}. We experience a significant increase ($+9\%$) with respect to baseline PSPNet trained on original clear images (\textit{Original}), and also outperform ($+2.1\%$) the finetuning with rain physics-based rendering~\cite{halder2019physics}. 
Both networks finetune \textit{Original} weights. The overall low numbers reported are impacted by the significant domain shift between Cityscapes and nuScenes.

\noindent\textbf{Disentanglement on heterogeneous datasets.}\label{sec:exp-fog-disentanglement} We now evaluate the effectiveness of the $\text{synth}\mapsto\text{WCS}_\text{fog}$ experiment which translates from synthetic Synthia to the real-augmented Weather CityScapes~\cite{halder2019physics} entangling fog of various intensities (from light to thick fog). 
Notice this task significantly differs from others for two reasons. 
First, unlike other experiments the model parameter -- the optical extinction coefficient, $\beta$ -- varies in the target dataset.
Second, the fog model is depending on the scene geometry~\cite{narasimhan2002vision}. 
This makes the disentanglement task non-trivial.
In our adversarial disentanglement, we however still regress a single $\beta = 28.61$ somehow averaging the ground truth values ($\beta \in [4, 40]$).

In Fig.~\ref{fig:qualitative-fog} results show we are able to generate images stylistically similar to target ones, but with geometrical consistency and varying $\beta$ (last 3 rows). Instead, MUNIT~\cite{huang2018multimodal} fails to preserve realism due to entanglement artifacts, visible in particular on elements at far (as buildings in the background). Please note that we intentionally do not show disentangled output for fairness, since the physical model always blocks the gradient propagation in the sky. More details on this will be discussed in Sec.~\ref{sec:discussion}. 
Randomizing $\beta \in [4, 40]$ we report GAN metrics results in Tab.~\ref{table:ganmetrics}, where the increased quality of images is quantified. LPIPS distance suffers from the absence of artifacts in our model-guided $\mathcal{T}_{W_{\text{Mod}}^{w_d}}(\tilde{w}_d)$, which artificially increases image variability. 
The physical model always renders correctly regions at far (e.g. the sky, which is always occluded), hence pure variability quantified by LPIPS is reduced (cf. above LPIPS definition).

\noindent\textbf{User study.} To further evaluate our disentanglement quality, we asked 56 users to rate images (details in Sec.~\ref{sec:methodology-user-study}). First, we presented our disentangled outputs and real images \textit{with} occlusions on the $\text{clear}{\mapsto}\text{rain}_{\text{drop}}$ and $\text{gray}{\mapsto}\text{color}_{\text{dirt}}$ tasks, where users were asked to rate for each image if "The camera lens is clean (no dirt, no raindrops)". 
Results in Fig.~\ref{fig:userstudy-sec2} show our strategy is better since the lens in our images is judge cleaner than target images. 
However, this does not assess if the underlying transformation (i.e. wetness or color) was properly learned.

Hence, secondly we compare translation realism with the MUNIT baseline, rating the statement ``The scene looks wet'' for $\text{clear}{\mapsto}\text{rain}_{\text{drop}}$ and ``The scene looks colorful'' for $\text{gray}{\mapsto}\text{color}_{\text{dirt}}$. We also include real source images (i.e. gray) in $\text{gray}{\mapsto}\text{color}_{\text{dirt}}$ to evaluate performances in the naive identity transformation, and target images in both to set upper bounds. \rev{Results in Figs.~\ref{fig:userstudy-sec5},~\ref{fig:userstudy-sec4}} clearly show the superiority of our approach with respect to the MUNIT, heavily reducing the gap with real target images. 

\rev{In a nutshell, the study demonstrates that disentanglement is fairly perceived by users (Fig.~\ref{fig:userstudy-sec2}) while preserving the learned underlying transformation (Figs.~\ref{fig:userstudy-sec5},~\ref{fig:userstudy-sec4})}.

\subsubsection{Neural-guided disentanglement}
\label{sec:exp-disentanglement-neural}
Referring to Tab.~\ref{tab:tasks-list} `Neural', we now evaluate our ability to disentangle visual traits with our neural guidance from Sec.~\ref{sec:method-supervised}, for Dirt disentanglement in the $\text{gray}\mapsto\text{color}_\text{dirt}$ task\rev{, by using available annotations instead of a physics-based prior.} %

\begin{figure*}[t]
	\centering
	\begin{subfigure}[b]{0.5\textwidth}
		\centering
		\tiny
		\setlength{\tabcolsep}{0.003\linewidth}
		\renewcommand{\arraystretch}{0.3}
		\begin{tabular}{ccccc}
			&&& \multicolumn{2}{c}{\textbf{Model-guided}}\\
			\cmidrule[1pt](){4-5}
			\includegraphics[width=.19\textwidth]{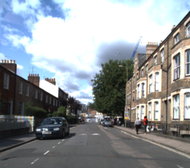} & \includegraphics[width=.19\textwidth]{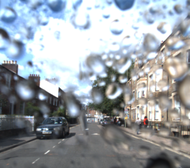} &
			\includegraphics[width=.19\textwidth]{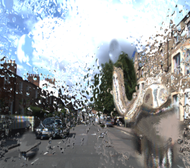} & \includegraphics[width=.19\textwidth]{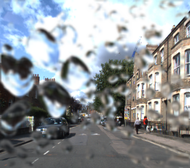} & \includegraphics[width=.19\textwidth]{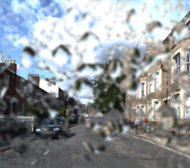}\\
			\\
			Source & Target & Porav et al.~\cite{porav2019can} & $\mathcal{T}_{W_{\text{Mod}}^{w_d}}(\tilde{w}_d)$ & $\mathcal{T}_{W_{\text{Mod}}}(\tilde{w})$
		\end{tabular}
		\caption{Sample images}\label{fig:porav-qualitative}
	\end{subfigure}  
	\begin{subfigure}[b]{0.24\textwidth}
		\centering
		\resizebox{1.0\linewidth}{!}{
			\small
			\setlength{\tabcolsep}{0.01\linewidth}
			\begin{tabular}{ccc}
				\toprule
				\textbf{Method} & \textbf{FID}$\downarrow$ & \textbf{LPIPS}$\downarrow$ \\
				\hline Porav et al. \cite{porav2019can} & 207.34 & 0.53  \\			
				Model-guided $\mathcal{T}_{W_{\text{Mod}}^{w_d}}(\tilde{w}_d)$ & \textbf{135.32} & 0.44 \\
				Model-guided $\mathcal{T}_{W_{\text{Mod}}}(\tilde{w})$ & 157.44 & \textbf{0.43} \\
				\bottomrule
			\end{tabular}
		}\vspace{1.4em}
		\caption{Benchmark on \cite{porav2019can}}
		\label{fig:porav-dist}
	\end{subfigure}
	\begin{subfigure}[b]{0.24\textwidth}
		\centering
		\includegraphics[width = 1.0\textwidth]{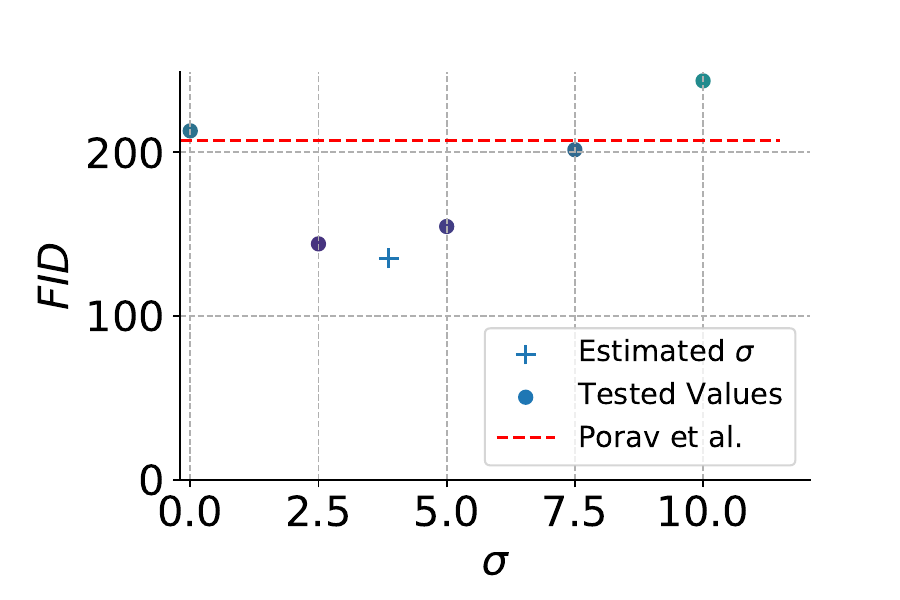}
		\caption{FID}
		\label{fig:porav-fid}
	\end{subfigure} %
	\caption{\textbf{Realism of the injected occlusion.} Our defocus blur $\sigma$ estimation grants an increased realism in raindrop rendering on the RobotCar \cite{porav2019can} dataset (\protect\subref{fig:porav-qualitative}), compared with Porav et al. \cite{porav2019can}. This is confirmed by quantitative metrics (\protect\subref{fig:porav-dist}). We report our model-guided translations using either differentiable parameter estimation only ($\mathcal{T}_{W_{\text{Mod}}^{w_d}}(\tilde{w}_d)$) or the full model parameter estimation ($\mathcal{T}_{W_{\text{Mod}}}(\tilde{w})$), outperforming Porav et al.~\cite{porav2019can} in both. In (\protect\subref{fig:porav-fid}), we evaluate the FID for different $\sigma$ values in [0, 10], showing that our regressed $\sigma$ value ($\sigma = 3.81$) actually leads to a local minimum.}
	\label{fig:porav}
\end{figure*}

\noindent\textbf{Evaluation.} We leverage here the WoodScape~\cite{yogamani2019woodscape} datasets having soiling semantic annotation as polygons. 
Following our training strategy (Fig.~\ref{fig:training-pipelines}, bottom), our neural guidance DirtyGAN~\cite{uricar2019let} (cf. Sec.~\ref{sec:exp-meth-neural}) is trained beforehand and frozen during the disentanglement.

The use of annotations boosts the overall quality and diversity, which is proved in Tab.~\ref{table:ganguided-ganmetrics} where our neural-guided outperforms both MUNIT baseline and our own model-guided version. 
Furthermore, since the ground truth for colorization is available, we evaluate in Tab.~\ref{table:ganguided-colorization} the effectiveness of disentanglement with SSIM and PSNR metrics (higher is better).
Here both disentanglement outperform MUNIT~\cite{huang2018multimodal} significantly, but model-guided is better. Arguably, we attribute this to the worse gradient propagation due to more occluded pixels with respect to our physical model\footnote{On average, DirtyGAN dirt covers 25.4\% of the image while our physical model covered 20.1\%. 
While this provides more realistic dirt masks (ground truth annotation is 29.6\%) we conjecture this leads to worse gradient propagation.}.

Finally, last 2 rows of Fig.~\ref{table:ganguided-qual} show our neural-guided strategy produces high quality \textit{colored} images \textit{without} occlusions (`Disentangled' row, $\mathcal{T}_{W_{\text{GAN}}}$) while injection of occlusions with optimal estimated parameters $\tilde{\theta}$ (`Target-style' row, $\mathcal{T}_{W_{\text{GAN}}}(\tilde{\theta})$) also mimics target appearance. In fact while both neural-guided disentanglement (Sec.~\ref{sec:exp-disentanglement-neural}) and physical model-guided disentanglement (Sec.~\ref{sec:exp-disentanglement-physics}) perform well, only our model-guided strategies controllability of the occlusion at inference. This is because of the explicit physical parameters in the models, that allows reinjecting unseen models at inference.

\subsection{Parameters estimation}\label{sec:parameter-estimation}
\label{sec:exp-validation}
We now evaluate the effectiveness of our parameter estimation for physical model-guided disentanglement, considering only differentiable parameters first and later extending to our full system. The neural-guided disentanglement strategy precludes this analysis due to the lack of explicit parameters.\\
\noindent\textbf{Differentiable model ($w=\{w_d\}$).}\label{exp-validation-diff} To evaluate realism, we leverage the RobotCar~\cite{porav2019can} dataset having pairs of clear/raindrop images.
Since there is no domain shift between image pairs, we set $G(x) = x$ and regress the defocus blur ($\sigma$) again following Sec.~\ref{sec:method-estimation}. 
The regressed $\sigma=3.87$ is used to render raindrops on clear images.
Using FID and LPIPS distances we measure perceived distance between real raindrop images and our model-guided raindrops translations ($\mathcal{T}_{W_{\text{Mod}}^{w_d}}(\tilde{w}_d)$) or the one of Porav et al.~\cite{porav2019can}. 
Fig.~\ref{fig:porav-dist} shows we greatly boost similarity\footnote{Please note that unlike previous experiments, here LPIPS is used for distance estimation (not diversity), so lower is better.} ($-72.02$ FID) with real raindrop images. This is qualitatively verified in Fig.~\ref{fig:porav-qualitative}, where our rendered raindrops are more similar to \textit{Target}. 
To provide insights about the quality of our minima, we also evaluate FID for arbitrary $\sigma$ values ($\sigma \in \{0.0, 2.5, 5.0, 7.5, 10\}$). Fig.~\ref{fig:porav-fid} proves that our estimated sigma best minimized perceptual distances despite the weak discriminator signal.

\begin{figure}[t]
	\centering
	\begin{subfigure}[m]{0.33\linewidth}
		\centering\tiny
		Defocus blur ($\sigma$)
		\includegraphics[width=\textwidth]{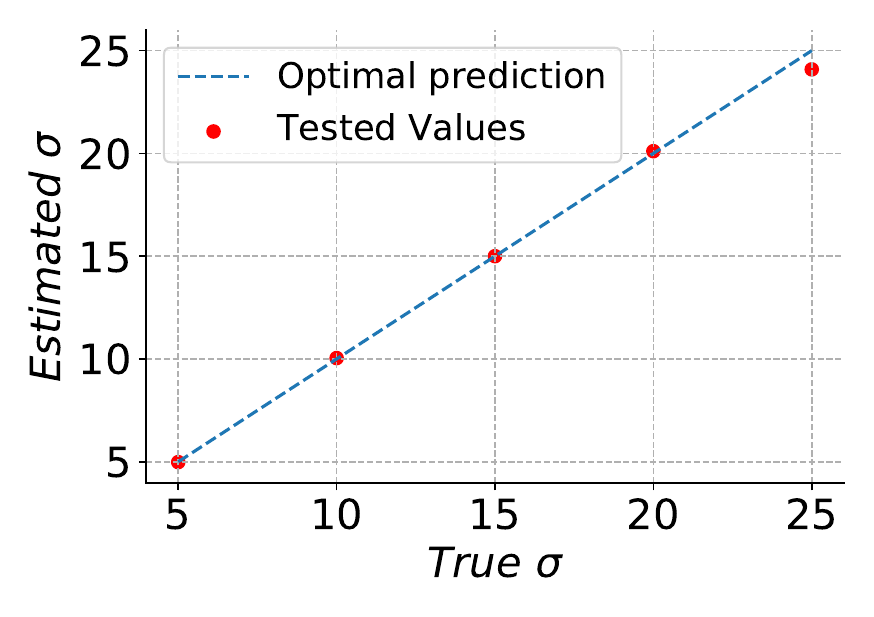}
		\caption{Raindrop}\label{plot:drops}
	\end{subfigure}\begin{subfigure}[m]{0.33\linewidth}
		\centering\tiny
		Transparency ($\alpha$)
		\includegraphics[width=\textwidth]{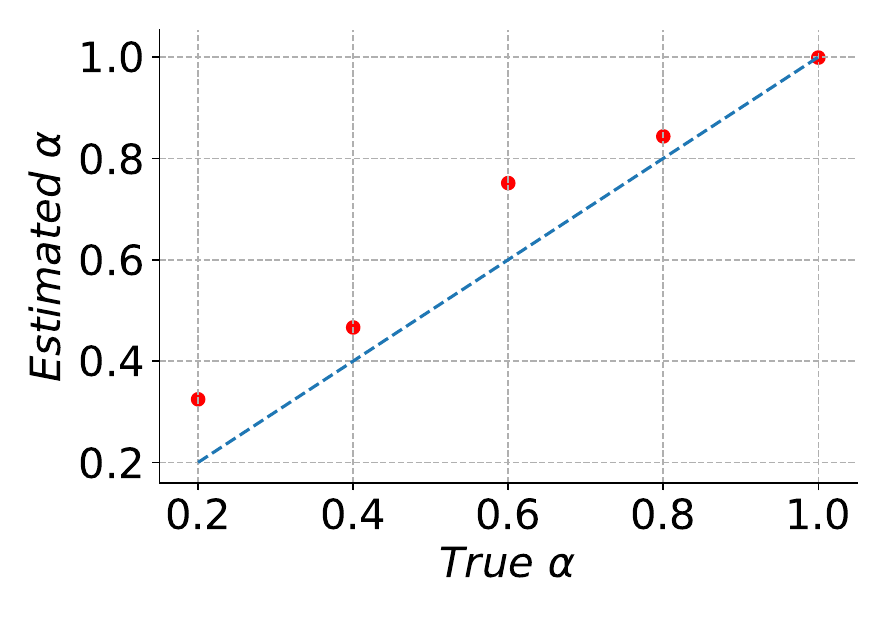}
		\caption{Dirt}\label{plot:dirt}
	\end{subfigure}\begin{subfigure}[c]{0.33\linewidth}
		\centering\tiny
		Optical extinction ($\beta$)
		\includegraphics[width=\textwidth]{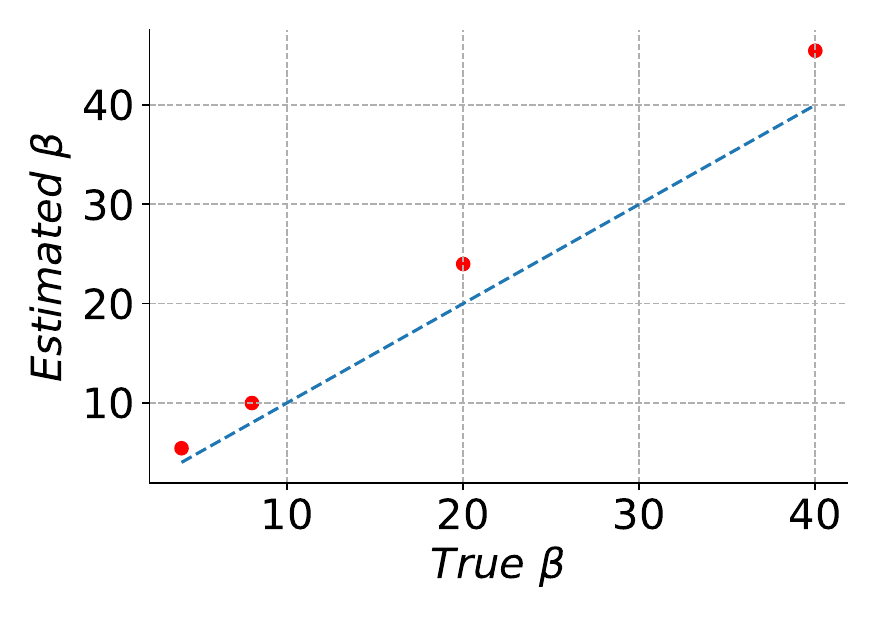}
		\caption{Fog}\label{plot:fog}
	\end{subfigure}
	\caption{\textbf{Evaluation of the model parameters regression.} The reliability of our parameter estimation is assessed on synthetic datasets augmented with arbitrary physical models acting as ground truth values. Comparing against our regressed value, our strategy performs better when low modifications on the estimated values corresponds to big visual changes (average error is 0.99\% for raindrops (\protect\subref{plot:drops}), 3.55\% for dirt (\protect\subref{plot:dirt})). For fog (\protect\subref{plot:fog}), we get an higher error of 23.51\% due to the low visual impact of high $\beta$ values.}
	\label{fig:plots}
\end{figure}

\begin{figure}[t]
	\centering
	\setlength{\tabcolsep}{0.005\linewidth}
	\renewcommand{\arraystretch}{0.3}
	\small
	\resizebox{1.0\linewidth}{!}{
		\begin{tabular}{cccc}
			\vspace{0.2em}
		    && \multicolumn{2}{c}{\textbf{Model-guided}}\\
			\cmidrule[1pt](){3-4}
			\includegraphics[width=0.24\linewidth]{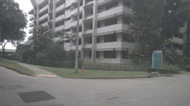}
			& \includegraphics[width=0.24\linewidth]{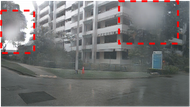}
			& \includegraphics[width=0.24\linewidth]{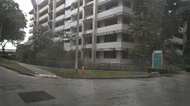}
			& \includegraphics[width=0.24\linewidth]{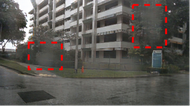}\\
			Source & MUNIT~\cite{huang2018multimodal} & $\mathcal{T}_{W_{\text{Mod}}^{w_d}}$ & $\mathcal{T}_{W_{\text{Mod}}}$
	\end{tabular}}
	\caption{\textbf{Full model on $\text{clear}\mapsto{}\text{rain}_\text{drop}$.}
		With complete parameter estimation ($\mathcal{T}_{W_{\text{Mod}}}$, rightmost), we achieve a slightly worse disentanglement than with manually-tuned non-differentiable parameters ($\mathcal{T}_{W_{\text{Mod}}}^{w_d}$), visible in red areas of $\mathcal{T}_{W_{\text{Mod}}}$. However, in both of our translations we generate typical rain traits as reflections with reasonable disentanglement, while baseline MUNIT~\cite{huang2018multimodal} has very evident raindrops entangled highlighted in red.}\label{fig:disentanglement-full}
\end{figure}

\begin{figure}[t]
    \centering
    \includegraphics[width=0.75\linewidth]{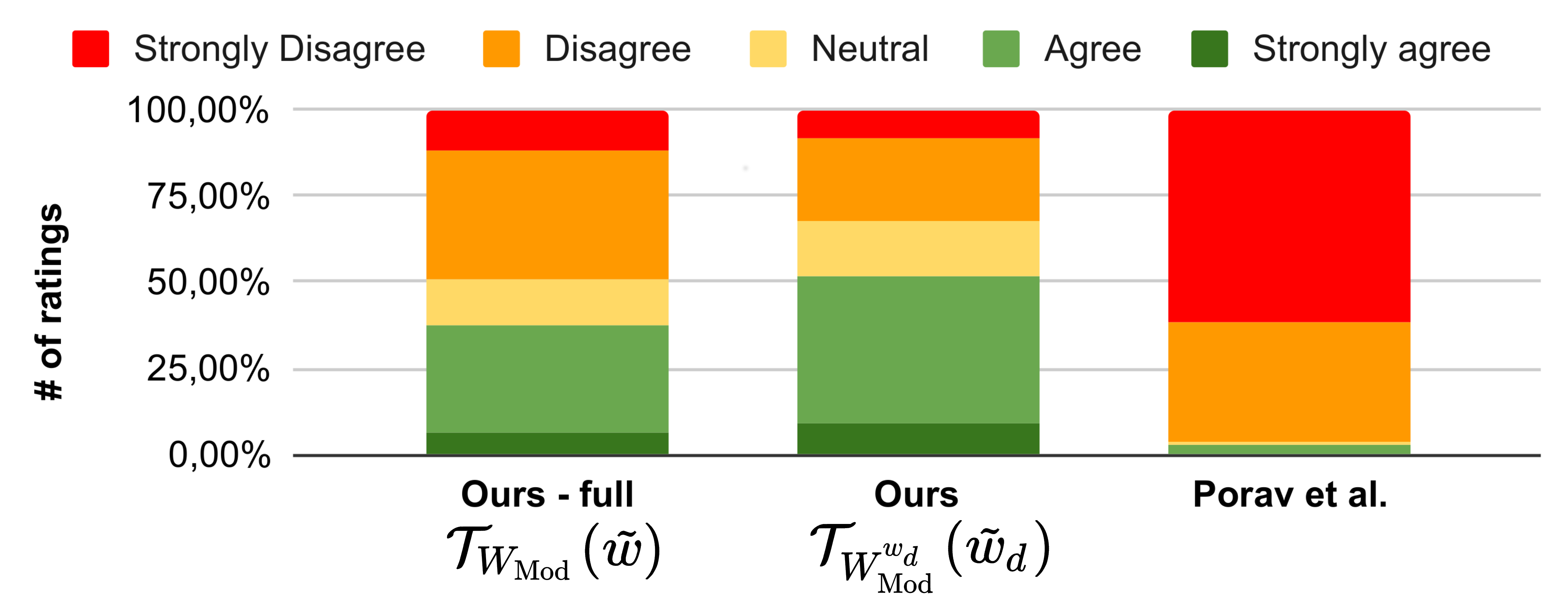}
    \caption{\textbf{Parameter estimation user study.} We presented users with \{\textit{Reference}, \textit{Model}\} image pairs where \textit{Reference} includes real drops and \textit{Model} has fake drops rendered with our method with differentiable only (Ours) or full (Ours - full) parameters estimation, or with Porav et al.~\cite{porav2019can}. Users were asked whether they agree on the statement "The drops of the Model resemble the drops of Reference". Thanks to our estimation strategy, we dramatically improve similarity to real raindrops.}     \label{fig:porav-userstudy}
\end{figure}

\label{sec:exp-validation-synthetic}
To measure the accuracy of our differentiable parameter regression (Sec.~\ref{sec:method-estimation}) we need paired images with and without physical traits with completely known physical parameters. 
To the best of our knowledge such dataset does not exists.
Instead, we augment RobotCar~\cite{porav2019can}, WoodScape~\cite{yogamani2019woodscape} and Synthia~\cite{ros2016synthia} with synthetic raindrops, dirt, and fog, respectively, with gradually increasing values of defocus blur ($\sigma$) for raindrop, transparency ($\alpha$) for dirt\footnote{In this experiment, we consider dirt with a fixed defocus blur value $\sigma$ and regress only $\alpha$ to increase the diversity of tasks.} and optical thickness ($\beta$) for fog. 
Using each augmented dataset, we then regress said parameters following Sec.~\ref{sec:method-estimation}.

Plots in Fig.~\ref{fig:plots} show estimation versus ground-truth. In average, the estimation error is $0.99\%$ for raindrop, $3.55\%$ for dirt, and $23.51\%$ for fog. 
The very low $\sigma$ error for raindrop is to be imputed to the defocus blur that drastically changes scene appearance, while higher error for $\beta$ must be imputed to the logarithmic dependency of the fog model. Nevertheless, translations preserve realism (cf. Fig.~\ref{fig:qualitative-fog}).\\

\noindent\textbf{Full model ($w=\{w_d,w_{nd}\}$).}\label{exp-validation-full} To evaluate the quality of our full raindrop model, we incorporate this time the non-differentiable parameters (i.e. $s$, $p$, $t$) which are estimated with our genetic strategy in Sec.~\ref{sec:method-genetic} for 4 types of drops, with a genetic population size of 10.
As shown in Fig.~\ref{fig:porav-dist}, LPIPS metric privileges our full model-guided estimation ($\mathcal{T}_{W_{\text{Mod}}}(\tilde{w})$) while FID suffers compared to using differentiable parameters only. 
However, we very significantly outperform~\cite{porav2019can} also qualitatively~(Fig.~\ref{fig:porav-qualitative}). 
The mitigated results are explained by the much more complex optimization problem having many more parameters, and by the limited computation time for genetic iterations. 
However, this let us foresee applications in high-dimensionality problems where manual approximation is not always possible or with a less accurate model (see ablations Sec.~\ref{sec:exp-ablation-complexity}). \rev{Moreover, we stress that the manual optimization could be challenging and time consuming~(cf.~Sec.~\ref{sec:exp-ablation}).}\\
Results on the $\text{clear}\mapsto{}\text{rain}_\text{drop}$ task in Fig.~\ref{fig:disentanglement-full} are coherent with above insights as the full model estimation, although effective, exhibits slightly lower quality disentanglement.

\noindent\textbf{User study.} We presented to users (see Sec.~\ref{sec:methodology-user-study}) couples of images with independent scenes in which the left one presented images with real drops taken from RobotCar~\cite{porav2019can}, while the right one included fake drops rendered with our model with differentiable only / all parameters estimated, or with Porav et al.~\cite{porav2019can}. Users were asked to compare raindrops appearance between the two images regardless of the represented scenes. From results shown in Fig.~\ref{fig:porav-userstudy}, it is evident that our method largely outperform the baseline in both configurations, indicating a higher quality of our raindrops also for the human preference metric.

\begin{figure}[t]
    \small
    \centering
    \begin{subfigure}{.35\linewidth}
	\centering
        \resizebox{1.0\linewidth}{!}{\renewcommand{\arraystretch}{1.1}\tiny
    	\setlength{\tabcolsep}{0.013\linewidth}
		\small
            \begin{tabular}{cccc}
            \toprule
            \textbf{Model}          & \textbf{IS$\uparrow$}    & \textbf{LPIPS$\uparrow$} & \textbf{CIS$\uparrow$}   \\\hline
            \textit{none} & 1.21 & 0.50 & 1.03 \\\hline
            Gaussian & 1.35 & 0.51 & 1.13 \\
            Refract & 1.46 & 0.50 & 1.12 \\
            Ours & \textbf{1.53} & \textbf{0.52} & \textbf{1.15} \\
            \bottomrule
        
            \end{tabular}}
    \caption{Model complexity.}
    \label{fig:ablation-complexity}
	\end{subfigure}\hfill
	\begin{subfigure}{0.60\linewidth}
	\centering
    {\renewcommand{\arraystretch}{0.2}
	\setlength{\tabcolsep}{0.006\textwidth}
    \resizebox{\linewidth}{!}{
    \begin{tabular}{cccccc}
 	     \multicolumn{2}{c}{\adjustbox{valign=m}{\rotatebox{90}{Source}}}&
 	     \adjustbox{valign=m}{\includegraphics[width=0.49\linewidth]{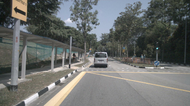}}
	     & \adjustbox{valign=m}{\includegraphics[width=0.49\linewidth]{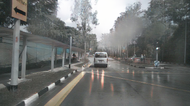}} & 
	     \adjustbox{valign=m}{\rotatebox{270}{\textit{(Baseline)}}} & \adjustbox{valign=m}{\rotatebox{270}{$\gamma = 0$}} \\
	     \adjustbox{valign=m}{\rotatebox{90}{$\gamma = 0.75$}} & \adjustbox{valign=m}{\rotatebox{90}{\textit{(Ours)}}} &
	     \adjustbox{valign=m}{\includegraphics[width=0.49\linewidth]{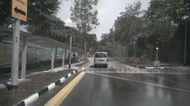}}
	     & \adjustbox{valign=m}{\includegraphics[width=0.49\linewidth]{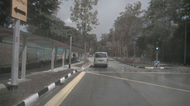}} & \multicolumn{2}{c}{	     \adjustbox{valign=m}{\rotatebox{270}{$\gamma = 1$}}} 
	     \\
         \end{tabular} }
         } %
    \caption{Disentanglement Guidance.}
    \label{fig:ablation-beta}
    \end{subfigure}
	\caption{\textbf{Ablations of model complexity and Disentanglement Guidance.} In (\protect\subref{fig:ablation-complexity}), we quantify disentanglement effects with simpler model having less variability (\textit{Refract}), or only color guidance (\textit{Gaussian}). Even if complexity is beneficial for disentanglement (\textit{Ours}), simple models permits disentanglement to some extent. In (\protect\subref{fig:ablation-beta}), we study the efficacy of the Disentanglement Guidance (DG) for different $\gamma$ values on $\text{clear}\mapsto{}\text{rain}_\text{drop}$ task. With $\gamma=0$ our approach fallbacks to the baseline and entangles occlusions, while with guidance $\gamma = 1$ the translation lacks important features such as  reflections and glares. With $\gamma=0.75$ we simultaneously avoid entanglements and preserve translation capabilities.}

\end{figure}

\begin{figure*}[t]
\revbox{
	\centering
	\setlength{\tabcolsep}{0.005\linewidth}
	\renewcommand{\arraystretch}{0.3}
	\small
	\resizebox{1.0\linewidth}{!}{
		\begin{tabular}{c|ccccc|c}
		Source & $\{\tilde{w}_d, w_{nd}^{\text{rand-1}}\}$ & $\{\tilde{w}_d, w_{nd}^{\text{rand-2}}\}$ & $\{\tilde{w}_d, w_{nd}^{\text{rand-3}}\}$ & $\{\tilde{w}_d, w_{nd}^{\text{rand-4}}\}$ & $\{\tilde{w}_d, w_{nd}^{\text{rand-5}}\}$ & \textbf{$\{\tilde{w}_d, \tilde{w}_{nd}\}$} \\
		\includegraphics[width=10em]{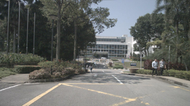} &
		\includegraphics[width=10em]{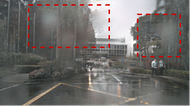} &
		\includegraphics[width=10em]{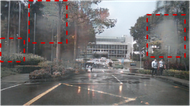} &
		\includegraphics[width=10em]{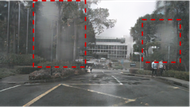} &
		\includegraphics[width=10em]{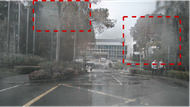} &
		\includegraphics[width=10em]{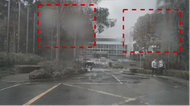} &
		\includegraphics[width=10em]{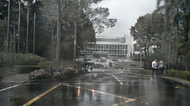} \\

		\end{tabular}}
\caption{\textbf{Benefit of genetic parameters optimization.} While genetic algorithms require to set optimization boundaries, even coarsely-defined boundaries can be used for achieving disentanglement. We sample 5 different random sets of parameters ($\{w_{nd}^{\text{rand-1}}, ..., w_{nd}^{\text{rand-5}}\}$) from boundaries set in minutes and combine them with $\tilde{w}_d$ estimation, achieving visible entanglement artifacts (highlighted with red boxes). 
Instead, using the same coarsely-defined boundaries for our genetic optimization our full parameter estimation $\{\tilde{w}, \tilde{w}_{nd}\}$ achieves reasonable disentanglement and qualitatively better results. 
Hence, our full optimization pipeline can benefit even from quick and coarse tuning of parameter boundaries.}\label{fig:ablation-random-wnd}

}

\end{figure*}

\begin{figure}[t]
\revbox{
	\centering
	\begin{subfigure}[b]{0.47\linewidth}
		\includegraphics[width=\linewidth]{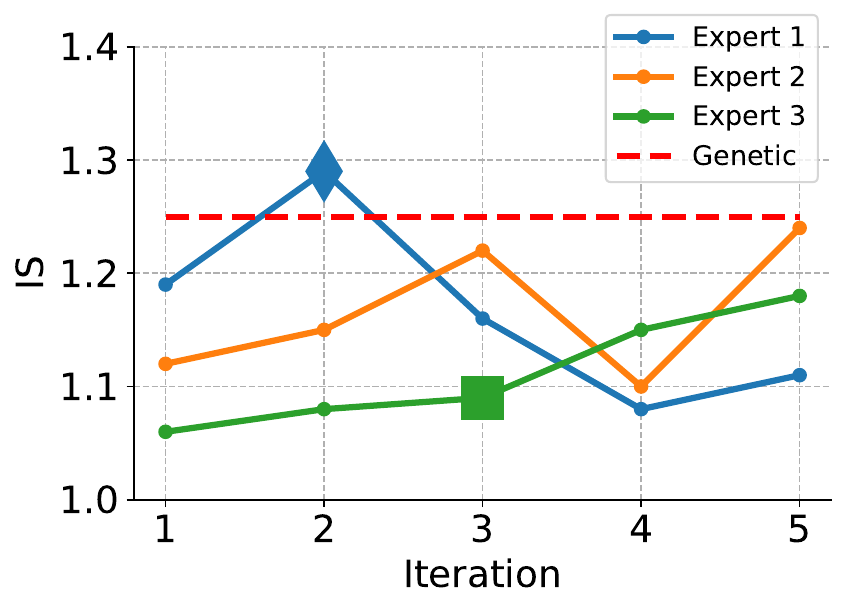}
		\caption{IS of trained models.}\label{fig:monkeys-1}
	\end{subfigure}%
	\hfill
	\begin{subfigure}[b]{0.52\linewidth}
		\resizebox{\linewidth}{!}{
		\begin{tabular}{cc}
		Source & Genetic\\
		\includegraphics[width=6em]{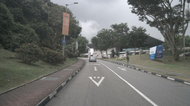} & \includegraphics[width=6em]{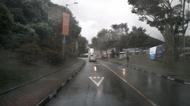}\\		
		\bluediamond~output & \greensquare~output \\	
		\includegraphics[width=6em]{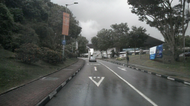} &			
		\includegraphics[width=6em]{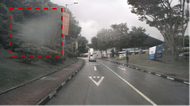}\\			
		\end{tabular}}
		\caption{Qualitative evaluation.} %
		\label{fig:monkeys-2}
	\end{subfigure}%
	\caption{\textbf{User-based VS Genetic-based optimization of $w_{nd}$.} In (\protect{\subref{fig:monkeys-1}}), we compare IS of user- or genetic-optimized $w_{nd}$. Expert computer vision users struggle to reach performances comparable to our genetic estimation due to the complexity of the parameter estimation task. In~(\protect{\subref{fig:monkeys-2}}), qualitative evaluation of the GAN output advocates that manual tuning can still lead to good performances (\bluediamond~model), but it can also lead to entanglement even after several iterations (\greensquare~model). Tuning $w_{nd}$ with our full pipeline (Genetic) prevents such failures, requiring also only one disentangled training.
	}\label{fig:monkeys}
}
\end{figure}

\subsection{Ablation studies}
\label{sec:exp-ablation}
We now ablate our proposal. We focus on the model-guided setting by tuning genetic processing, altering model complexity, changing models, or removing disentanglement guidance. \rev{We also further investigate  and compare different training strategies.}

\noindent\textbf{Model complexity.} \label{sec:exp-ablation-complexity} We study the influence of the model on disentanglement for the $\text{clear}\mapsto{}\text{rain}_\text{drop}$ on nuScenes~\cite{caesar2019nuscenes} task. Specifically, we evaluate three raindrop models of decreasing complexity: 1) Our model from Sec.~\ref{sec:exp-meth-model} (named \textit{Ours}). 2) The same model but without shape and thickness variability (\textit{Refract}), and 3) A naive non-parametric colored Gaussian-shape model (\textit{Gaussian}). 
Note that \textit{Gaussian} is deprived of any refractive property as it uses fixed color, and does not regress any physical parameters. In Fig.~\ref{fig:ablation-complexity}, we report GAN metrics for all models following Sec.~\ref{sec:exp-disentanglement-physics}. Even if increasing model complexity is beneficial for disentanglement, very simple models still lead to a performance boost. We advocate the best performances of \textit{Ours} to a more effective discriminator fooling during training, as consequence of increased realism.

\noindent\textbf{Model choice.} \label{sec:exp-ablation-choice} To also evaluate whether injected features only behave as adversarial noise regardless of the chosen model, we trained on RobotCar~\cite{porav2019can} (as in Sec.~\ref{sec:exp-validation}) though purposely using an incorrect model as watermark, dirt, fence.
Evaluating the FID against real raindrop images, we measure $\mathbf{135.32}$ (raindrop) / $329.17$ (watermark) / $334.76$ (dirt) / $948.71$ (fence), proving necessity of using the ad-hoc model.

\noindent\textbf{Disentanglement Guidance (DG).} \label{sec:exp-ablation-dg} We use the nuScenes $\text{clear}\mapsto{}\text{rain}_\text{drop}$ task to visualize the effects of different DG strategies (Sec.~\ref{sec:method-dg}). For varying values of the DG threshold $\gamma$ in Fig.~\ref{fig:ablation-beta} we see results ranging from no guidance ($\gamma = 0$) to strict guidance ($\gamma = 1$). With lax guidance ($\gamma = 0$), we fall back in the baseline scenario with visible entanglement effects, while with $\gamma = 1$ we do achieve disentanglement, at the cost of losing important visual features as reflections on the road. Only appropriate guidance ($\gamma = 0.75$) achieves disentanglement and preserves realism.

\subsubsection{Full model}
\noindent\textbf{Non-differentiable genetic estimation.} We study the effectiveness of our genetic estimation ablating the population size of our raindrop model on RobotCar~\cite{porav2019can} as in Sec.~\ref{sec:exp-validation}. We test our algorithm with population size~10/25/50/100, obtaining FID 157.44/153.32/151.21/\textbf{149.09} and LPIPS \textbf{0.43}/0.44/0.44/\textbf{0.43}. While we observe an obvious increase in performances, this comes with additional computation times, hence we used the lowest population size of 10 for all tests. Nevertheless, this opens doors to potential improvements in the full parametric estimation.

\noindent\rev{\textbf{Non-differentiable boundaries of $w_{nd}$.} Genetic algorithms requires optimization boundaries for each parameter (i.e. the \textit{min} and the \textit{max} of each parameter), so one could argue that $w_{nd}$ still requires manual tuning, therefore lowering the interest of our full estimation pipeline. 
However, our empirical studies demonstrate that parameter boundaries only takes a few minutes, while precise manual tuning required for differentiable-only optimization~(Sec.~\ref{sec:method-estimation}) takes days as it requires multiple training.
In an effort to provide evidence of the coarse boundaries definition, we randomly sampled 5 sets of $w_{nd}$ within said boundaries and report disentanglement results in Fig.~\ref{fig:ablation-random-wnd}.
Simply sampling parameters within the boundaries (center) achieve far less good disentanglement w.r.t. our full estimation pipeline (right).}

\subsubsection{Differentiable-only model}
\noindent\rev{\textbf{Manual estimation of $w_{nd}$}. To provide further proof on the interest of optimizing $w_{nd}$ with genetic algorithms, we perform an additional user study with three computer vision experts. To each expert, we show real rainy images of $\text{clear}\mapsto\text{rain}_{\text{drop}}$~(see~Tab.~\ref{tab:tasks-list}) and ask the latter to manually tune $w_{nd}$ of the drop model to reproduce the target drop appearance. 
	We then estimate the optimal remaining differentiable parameters $\tilde{w}_d$ and train a disentangled network, showing to the same expert the qualitative results obtained with the tuned parameters. 
	We finally asked to update the manually estimated values to improve disentanglement. 
	We perform multiple iterations and quantify performances in terms of Inception Score~(IS). 
	In Fig.~\ref{fig:monkeys}, it is visible how users difficultly improve performances even after 5 iterations, while with the full estimation pipeline we boost results with no manual tuning. 
	Altogether, this demonstrates how using our full pipeline can ease the estimation task and save computational time.
}

\section{Discussion}
\label{sec:discussion}
To our best knowledge, we have designed the first unsupervised strategy to disentangle physics-based features in i2i. 
Good qualitative and quantitative performances showcase promising interest for several applications, still there are peculiar points and limitations which we now discuss.

\begin{table}[t]
\revbox{
\scriptsize
\centering
\setlength{\tabcolsep}{0.01\linewidth}
\renewcommand{\arraystretch}{1.0}
\resizebox{\linewidth}{!}{%
\begin{tabular}{cc|c|cccc}
	\toprule
	\multirow{2}{*}{\textbf{Guidance}} & \multirow{2}{*}{\textbf{Param. estimation}} & \multirow{2}{*}{\textbf{Editable}} & \multicolumn{4}{c}{\textbf{Requirements}}\\
	& & & Annotations & Ad-hoc GAN & Model design & Manual $w_{nd}$ tuning\\
	\midrule
	Neural & None & & \checkmark & \checkmark & & \\
	Model & $w_d$ & \checkmark & & & \checkmark & \checkmark \\
	Model & $\{w_d, w_{nd}\}$ & \checkmark & & & \checkmark & \\
	\bottomrule
\end{tabular}
}%
\caption{\textbf{Comparison of the disentanglement strategies.} Model-guided strategies do not require annotations and ad-hoc generative networks, but they rely on the availability of a somehow realistic physics model. When using neural-guided disentanglement, the ability to modify physical parameters of the model (``Editable'') is lost. We overcome the need of cumbersome manual tuning of $w_{nd}$ with genetic optimization in our full strategy. However, results in Sec.~\ref{sec:exp} advocate that best disentanglement performances are still obtained by manually sizing each non differentiable parameter, at the cost of intensive labor and many trainings.}\label{tab:discussion-comparison}
}
\end{table}

\noindent\rev{\textbf{Comparison of different disentanglement strategies.}}
\rev{We propose three different disentanglement strategies.
In Tab.~\ref{tab:discussion-comparison}, we compare them, highlighting advantages and disadvantages that could be crucial for choosing a disentanglement strategies in an applicative scenario. 
While the differentiable-only estimation strategy performs best in terms of disentanglement, it is also time consuming due to manual tuning of $w_{nd}$. The applicability of neural-guided disentanglement depends on annotations availability, and prevents outputs editing capabilities at inference. 
Ultimately, one should prefer model-guided disentanglement if a model is accessible.
}

\noindent\textbf{Independence assumption.}
For unsupervised disentanglement, we assume the physical model to be completely independent from the scene, in order to use our intuition about marginal separation (see Sec. ~\ref{sec:method-disentanglement} and Eq.~\ref{eq:joint}). However, since physical models may need the underlying scene to correctly render desired traits, one may argue their appearance is not completely disentangled. While this is true from a visual point of view, it is not from a physical one. Let's interpret disentanglement properties to be dependent on scene \textit{elements}. In presence of disentanglement, the same physical model could be applied to different objects regardless of what they are. For instance, we could use the same raindrop refraction map on either roads or buildings with identical parameters. In this sense, $G(x)$ dependency in physical models is not impacting our visual independence assumption.

\noindent\textbf{On partial entanglement issues.}
We observe in some cases that gradient propagation can be affected by fixed entanglement of occlusion features. This is the case for example for sky regions in fog (Sec.~\ref{sec:exp-fog-disentanglement}) because physics~\cite{narasimhan2002vision} formalizes that regardless of its intensity fog is always entangled at far. 
In such scenarios, disentanglement will perform poorly because the generator will not get any discriminative feedback.
In many other cases however, Disentanglement Guidance (DG, Sec.~\ref{sec:method-dg}) mitigates the phenomenon as it blocks injection of the physical model in relevant image regions. 
We conjecture that the effectiveness could be extended by varying DG at training time to ensure a balanced gradient propagation.

\noindent\textbf{On genetic estimation effectiveness.}
The sub-optimal performances of our genetic estimation of $w_{nd}$ are imputed to the much more complex search space, in which we vary all parameters of our physical model simultaneously. \rev{It is worth noting that manually tuning non-differentiable parameters requires many trainings, while relying on genetic optimization achieves acceptable results in a single complete training.} Also, we did set fairly large search boundaries for $w_{nd}$ \rev{(as evaluated in Sec.~\ref{sec:exp-ablation})}, but one could envisage a mixed training in which the search space is limited to reasonable hand-tuned boundaries. In this sense, genetic estimation of $w_{nd}$ could be seen as a minimum mining technique, ensuring increased performances on the hand-tuned values.

\noindent\textbf{Acknowledgments.}
This work used HPC resources from GENCI–IDRIS (Grants 2020-AD011012040 and 2021-AD011012808). This work was partially funded by French project SIGHT (ANR-20-CE23-0016).

\bibliographystyle{IEEEtran}
\bibliography{egbib}

\vspace{-40em}
\begin{IEEEbiography}[{\includegraphics[width=1in,height=1.10in,clip,keepaspectratio]{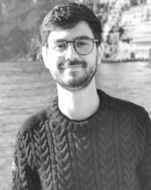}}]{Fabio Pizzati} completed his PhD at Inria Paris in 2022, supervised by Raoul de Charette. Before that, he received his MSc in Computer Engineering in 2018 and his BSc in Computer, Communication and Electronic Engineering in 2015, both from University of Parma, Italy. His research focuses on generative networks, vision in adverse weather and domain adaptation.
\end{IEEEbiography}
\vspace{-40em}
\begin{IEEEbiography}[{\includegraphics[width=1in,height=1.10in,clip,keepaspectratio]{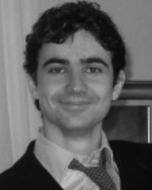}}]{Pietro Cerri} received the Dr. Eng. degree in computer engineering from the Università di Pavia, Pavia, Italy, in 2003, and the Ph.D. degree in information technologies from the Università di Parma, Parma, Italy, in 2007. He is currently team leader at Vislab, an Ambarella Company. His research is mainly focused on computer~vision and sensors fusion approaches for the development of advanced driver assistance systems.
\end{IEEEbiography}
\vspace{-40em}
\begin{IEEEbiography}[{\includegraphics[width=1in,height=1.10in,clip,keepaspectratio]{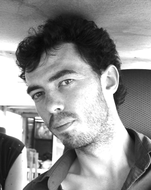}}]{Raoul de Charette} is a researcher at Inria Paris since 2015. He received the MSc degree in arts and computer graphics from Paris 8 Uni. and the PhD degree in computer vision for autonomous driving from Mines ParisTech in 2012. He was with Mines ParisTech (2008-2011, 2012-2013), Carnegie Mellon University (2011), and University of Makedonia (2014). He now leads the \mbox{Astra-vision} group in the Astra team, Inria. 
\end{IEEEbiography}

\end{document}